\documentclass[english]{article}
\usepackage[T1]{fontenc}
\usepackage[latin9]{inputenc}
\usepackage{geometry}
\usepackage{multirow}
\usepackage{amsmath}
\usepackage{amsthm}
\usepackage{amssymb}
\usepackage{graphicx}
\usepackage{setspace}
\usepackage[authoryear]{natbib}
\makeatletter

\providecommand{\tabularnewline}{\\}

\theoremstyle{plain}
\newtheorem{thm}{\protect\theoremname}
\theoremstyle{plain}
\newtheorem{prop}[thm]{\protect\propositionname}

\usepackage{hyperref}
\usepackage[lined,boxed]{algorithm2e}
\usepackage{cleveref}
\usepackage{tikz}
\usetikzlibrary{fit,positioning,arrows,automata,calc, bayesnet}
\author{}
\date{}

\@ifundefined{showcaptionsetup}{}{%
 \PassOptionsToPackage{caption=false}{subfig}}
\usepackage{subfig}
\makeatother

\usepackage{babel}
\providecommand{\propositionname}{Proposition}
\providecommand{\theoremname}{Theorem}

\begin{document}
\title{Deep Multistage Multi-Task Learning for Quality Prediction of Multistage
Manufacturing Systems }
\author{Hao Yan$^{1}$, Nurretin Dorukhan Sergin$^{1}$, and William A. Brenneman$^{2}$,
\\
Stephen Joseph Lange$^{3}$, and Shan Ba$^{4}$\\
 $^{1}$School of Computing, Informatics, and Decision Systems Engineering,
\\
Arizona State University, Tempe, Arizona \\
 $^{2}$The Procter \& Gamble Company, Cincinnati, Ohio\\
$^{3}$ProcessDev, LLC, Cincinnati , OH\\
$^{4}$LinkedIn Corporation }
\maketitle
\begin{abstract}
In multistage manufacturing systems, modeling multiple quality indices
based on the process sensing variables is important. However, the
classic modeling technique predicts each quality variable one at a
time, which fails to consider the correlation within or between stages.
We propose a deep multistage multi-task learning framework to jointly
predict all output sensing variables in a unified end-to-end learning
framework according to the sequential system architecture in the MMS.
Our numerical studies and real case study have shown that the new
model has a superior performance compared to many benchmark methods
as well as great interpretability through developed variable selection
techniques. 

\vspace{0.2cm}

{\textbf{Keywords:} deep neural network, multi-task learning, multistage
manufacturing process, quality prediction} 
\end{abstract}

\section{Introduction \label{sec: Introduction}}

A Multistage Manufacturing System (MMS) refers to a system consisting
of multiple stages (e.g., units, stations, or operations) to fabricate
a final product. In most cases, an MMS has a sequential configuration
that links all stages with a directed flow from the initial part/material
to the final product. In an MMS, massive data are collected from the
in-situ multivariate sensing of process variables and the product
quality measurements from intermediate stages and the final stage.
It is challenging to model the MMS data due to the following characteristics:
1) Multiple correlated output variables: The MMS can have hundreds
of output sensing variables at different stages that measure different
quality attributes of the product in different stages. These quality
responses are often correlated and evolve together with the product
in the MMS. 2) High dimensional input variables: In an MMS, there
can be hundreds of input variables (e.g., process sensors) in each
stage as well as a large number of stages/stations. 3) Complex sequential
dependency between stages and sensor measurements: The product quality
from an MMS is determined by complex interactions among multiple stages.
For example, the quality characteristics of one stage are not only
influenced by local process variations within that stage, but also
by the variation propagated from upstream stages. 

The methodology developed in this paper is motivated by a diaper manufacturing
system, which is composed of several sequential converting process
steps with associated equipment components, which are controlled through
many process factors. The setpoints for these factors and their variation
may affect the reliability of the process and the quality of the product
produced by it. Performance data of the converting equipment, as well
as in-process product quality measures, are collected and stored by
sensors and high-speed cameras installed on the production line. The
time-series data may be high-frequency data (sampled every several
milliseconds over seconds or minutes) and/or low-frequency data (samples
collected on a one-minute frequency over months). The number of variables
in this MMS is in the range of 500-1000. This data is used to detect
abnormal performance early enough to prevent machine stops, quality
degradation or scrap from rejected defective products. 
To model the variation propagation of the entire MMS, the stream of
variation (SoV) theory was proposed to model the variation propagation
in an MMS \citep{Shi2006}. For example, when one of the stages of
the multistage system experiences a malfunction, SoV can be used to
model the consequence of such a change on the final product \citep{Li2009,Lawless1999,Jin2009}.
However, in modern manufacturing processes, such as semiconductor
manufacturing or diaper manufacturing process, it is often not easy
to define the system state explicitly due to the complex system mechanism.
In conclusion, since classic approaches assume the design information
and physical law of the process to be perfectly known,  these classical
approaches cannot be used for complex systems with unknown engineering
or science knowledge.

In industrial practice and literature, to deal with more complicated
types of data and nonlinear dependencies, predictive modeling techniques
are often used. However, these models are typically designed based
on a single output variable such as decision tree \citep{bakir2006defect,jemwa2005improving}
or neural network \citep{zhou2006hierarchical,tam2004diagnosis,chang2002assessing}
and have the following major limitations: 1) Due to the independent
modeling of each output variable at each stage, the correlation between
each output variable is not considered. 2) Due to the need to have
one model for every single variable, it lacks a unified way of modeling
the MMS and variation propagation throughout the system. Hundreds
or even thousands of independent models are needed, which is very
hard to train and deploy. In addition, this greatly increases the
number of parameters in modeling the entire MMS, which could lead
to severe over-fitting when the number of samples is limited. 3) Due
to the need for a unified model for process monitoring and control,
there may be trade-offs between different output variables (e.g.,
quality variables). A joint modeling framework can be used for process
optimization of multiple input variables, considering the trade-offs
between different output variables simultaneously.

To address the aforementioned challenges, we aim to develop a data-driven
approach, namely the deep multistage multi-task learning (DMMTL) framework,
to link all the input and output variables in a sequential MMS. To
the best of the authors' knowledge, this is the first \textit{end-to-end
system-level predictive modeling framework} in modeling all input
and output variables jointly in an MMS. The proposed framework uses
a latent state representation similar to the SoV model, but instead
of specifying the hidden state representation manually, the proposed
framework aims to learn the latent state representation and how it
propagates through the MMS in an end-to-end fashion through all the
output variables simultaneously. We further make many improvements
on the model architecture to model the MMS, such as \textit{an independent
state transition model} and the group lasso penalty. Finally, since
model interpretability and diagnostics are also important, we will
also demonstrate how the proposed DMMTL is able to rank the most important
input variables according to each output variable for system diagnostics.

To model multiple output variables simultaneously, the proposed DMMTL
is also inspired by the recent development in multi-task learning
(MTL). MTL is a sub-field of statistical machine learning in which
multiple correlated learning tasks are solved at the same time while
exploiting the commonalities across tasks while modeling their differences
\citep{caruana1997multitask}. The proposed DMMTL can benefit through
incorporating the use of MTL to jointly model all output variables
from all stages simultaneously since these output variables often
measure correlated attributes of the same product in the MMS. However,
MTL by itself can only be used for joint modeling of multiple sensing
variables within the same stage, which fails to model the out variables
with sequential order. 

Finally, we present an overview of the proposed methodology compared
to existing manufacturing methodology such as stream of variation
(denoted as ``SoV'') \citep{Shi2006}, predictive modeling (denoted
as ``Predictive'') \citep{kuhn2013applied}, and multi-task learning
(denoted as ``MTL'') \citep{obozinski2006multi} in \Cref{table:
CompLiterature}. The proposed method is the only one that can handle
the unknown physical state (e.g., unknown transition functions), nonlinear
models, variable selection, joint multistage modeling, and joint multiple
sensor modeling within the stages. Finally, the proposed method can
also achieve diagnostics and interpretability by using the gradient
tracking techniques developed in this paper.

\begin{table}
\centering \caption{Comparison of Literature}
\begin{tabular}{|l|l|l|l|l|}
\hline 
 & Proposed & SoV & Predictive & MTL\tabularnewline
\hline 
Unknown Physical State Definition & $\checkmark$ & $\times$ & $\checkmark$ & $\checkmark$\tabularnewline
\hline 
Nonlinear Extension & $\checkmark$ & $\times$ & $\checkmark$ & $\times$\tabularnewline
\hline 
Variable Selection & $\checkmark$ & $\times$ & $\checkmark$ & $\checkmark$\tabularnewline
\hline 
Sequential Multistage Modeling & $\checkmark$ & $\checkmark$ & $\times$ & $\times$\tabularnewline
\hline 
\begin{tabular}{@{}l@{}}
Joint Multi-sensor Modeling\tabularnewline
Within Stages\tabularnewline
\end{tabular} & $\checkmark$ & $\checkmark$ & $\times$ & $\times$\tabularnewline
\hline 
Diagnostics & $\checkmark$ & $\checkmark$ & $\times$ & $\times$\tabularnewline
\hline 
\end{tabular}\label{table: CompLiterature}
\end{table}

The remaining parts of the paper are organized as follows. \Cref{sec:
Literature-Review} reviews the related literature in manufacturing
modeling. In \Cref{sec: Methodology-Development}, we develop the
proposed methodology for quality prediction and variable selection.
\Cref{sec: EstimationDiagnosis} discusses the estimation procedure
and variable selection techniques. \Cref{sec:Simulation} presents
a simulation study to compare the proposed framework with several
traditional methods in the literature. In \Cref{sec:Case-Study},
we apply the proposed framework to model a diaper manufacturing line.
In \Cref{sec:Conclusion}, we provide concluding remarks and future
direction. For a more detailed comparison of the proposed method with
SoV, please refer to \Cref{sec: sov}. 

\section{Literature Review \label{sec: Literature-Review}}

In this section, we will review the related literature in the modeling
of manufacturing systems. We briefly classify existing techniques
into two types, single-stage modeling and multistage modeling.

Single-stage models typically focus on process monitoring \citep{joe2003statistical,Kourti1996,MacGregor1995}
and the prediction \citep{Hao2016,Jin2007} of sensing variables observed
in the same stage of the manufacturing system. For process monitoring,
uni-variate \citep{shewhart1931economic}, multivariate \citep{lowry1995review},
profile-based \citep{woodall2007current}, multi-channel-profile-based
\citep{paynabar2013monitoring,Zhang2018}, and image-based \citep{Yan2015,Yan2017}
process monitoring techniques are developed. For quality prediction,
regression and classification techniques such as linear regression
\citep{skinner2002multivariate}, logistic regression \citep{Jin2007},
tensor learning \citep{yan2018structured}, decision trees \citep{bakir2006defect,jemwa2005improving},
and neural network \citep{zhou2006hierarchical,tam2004diagnosis,chang2002assessing}
are applied to relate the input variables (e.g. process sensing variables)
with the output variables (e.g. quality sensing variables). Despite
the use of nonlinear methods and the ability to incorporate heterogeneous
high-dimensional data, it still lacks a unified framework for modeling
the variation propagation among stages.

To model multistage systems, SoV has been successfully implemented
in the multistage automotive assembly process \citep{Jin1999a,Apley1998,Ceglarek1996,Ding2005,Shiu1997}
and machining process \citep{Huang2002,Liu2009,Zhou2003,Abellan-Nebot2012}.
For example, SoV introduces the state space representation to quantify
the system status. In a traditional SoV model, the state variables
are defined physically (e.g., geometry deviation of the product \citep{Ding2005}).
For a more detailed literature review on MMS models, please refer
to \citep{Shi2006}. There are some other techniques besides SoV that
have been developed for multistage modeling. Bayesian network techniques
\citep{Friedman1997,Jensen1996} have been proposed to model the complex
dependency between multiple manufacturing stages for both process
monitoring \citep{Liu2014,yu2013novel,liu2013objective} and quality
prediction \citep{reckhow1999water,correa2008bayesian}. However,
these techniques assume that the complex dependencies between multiple
output variables are known and require the feature selection techniques
to be used beforehand, and thus cannot be applied to a system with
unknown transition and dependency. In the literature, reconfigured
piece-wise linear regression trees are developed \citep{Jin2012}
to take advantage of intermediate quality variables and model the
nonlinear relationship between the sequential order of the input and
output variables. However, this technique cannot perform feature selection
for a large number of sensors and it also assumes the same quality
responses are measured in the intermediate stages. Furthermore, it
optimizes the model in a greedy stage-wise approach, which may suffer
from local optimality. In conclusion, similar to SoV, most of the
techniques in the literature focus on either manual selection of important
sensors, manual extraction of useful features transformation, and
clear definition of system state and transition before the MMS modeling
can be applied. But these techniques cannot be used for quality prediction
of complex systems with unknown architectures \citep{Ding2005,Apley1998,Zhou2003,Zhou2004,Ceglarek1996}.

\section{Methodology Development \label{sec: Methodology-Development}}

In this section, we will first define the problem setting and notations
in \Cref{subsec: notation} followed by our proposed DMMTL framework
in \Cref{subsec:The-Proposed-MMS}. We further derived a more efficient
optimization algorithm to handle the non-smooth loss function and
penalty in \Cref{subsec:Optimization-Algorithm}. 

\subsection{Problem Setting and Notation \label{subsec: notation}}

We denote $\mathbf{x}_{k}=(\mathbf{x}_{k,1},\cdots,\mathbf{x}_{k,n_{x,k}})^{T}$
is a vector of input variables (i.e. process sensing variables) in
stage $k$, where $x_{k,i}$ is the $i^{th}$ sensing variable in
stage $k$, and $n_{x,k}$ is the total number of input sensing variables
in stage $k=1,\cdots,K$, where $K$ is the number of stages. We denote
$\mathbf{y}_{k}=(y_{k,1},\cdots,y_{k,n_{y,k}})^{T}$ as the output
variables (i.e. product quality sensing variables) in stage $k$,
where $y_{k,j}$ is the $j^{th}$ sensing variable in stage $k$ and
$n_{y,k}$ is the total number of output sensing variables in stage
$k$. To link multiple stages together, we introduce the hidden state
variable $\mathbf{h}_{k}$, which is a vector to represent the state
of the product in stage $k$. For simplicity, we assume the hidden
state variable $\mathbf{h}_{k}$ is of the same dimension $n_{h}$
across different states. The goal of this research is to build a multi-task
learning framework to predict the quality indices $\mathcal{Y}=\{\mathbf{y}_{1},\cdots,\mathbf{y}_{K}\}$
measured at different stages given $\mathcal{X}=\{\mathbf{x}_{1},\cdots,\mathbf{x}_{K}\}$.
In this section, we assume that we are only dealing with data with
sample size $1$. We will discuss how to extend this to the mini-batch
version utilizing mini-batch stochastic gradient descent with multiple
samples in Appendix D in detail.

\subsection{Proposed Framework \label{subsec:The-Proposed-MMS}}

We will introduce our proposed DMMTL model to solve the aforementioned
challenges by learning the hidden state representation $\mathbf{h}_{k}$
from data. Here, $\mathbf{h}_{k}$ should contain the information
not only to predict the current state output $\mathbf{y}_{k}$ but
also the future stage output $\mathbf{y}_{k'}$ with $k'>k$. In $k^{th}$
manufacturing stage, we will define the transition function to model
the state transition between $\mathbf{h}_{k-1}$ and $\mathbf{h}_{k}$
and the emission function to model the relationship between $\mathbf{h}_{k}$
and $\mathbf{y}_{k}$ as learnable parametric functions with model
parameters $\boldsymbol{\theta}_{k}^{h},\boldsymbol{\theta}_{k}^{g}$
as 
\begin{equation}
\mathbf{h}_{k}=f_{k}(\mathbf{h}_{k-1},\mathbf{x}_{k};\boldsymbol{\theta}_{k}^{h}),\mathbf{y}_{k}=g_{k}(\mathbf{h}_{k};\boldsymbol{\theta}_{k}^{g})+\mathbf{\epsilon}_{yk},\label{eq: generalrecursive}
\end{equation}
$\mathbf{\epsilon}_{yk}\sim N(0,\sigma^{2})$. One example of such
architecture is listed in (\ref{eq: RNN}), which we use a one-layer
neural network to model the nonlinear state transition and emission
function. 
\begin{equation}
\mathbf{h}_{k}=\sigma(\mathbf{W}_{xk}\mathbf{x}_{k}+\mathbf{U}_{hk}\mathbf{h}_{k-1}+\mathbf{b}_{hk}),\mathbf{y}_{k}=\mathbf{V}_{yk}\mathbf{h}_{k}+\mathbf{b}_{gk}+\mathbf{\epsilon}_{yk},\label{eq: RNN}
\end{equation}
where $\sigma(x)=1/(1+\exp(-x))$ is the activation function. Define
the stage transition model parameters as $\boldsymbol{\theta}_{k}^{h}=\{\mathbf{U}_{hk},\mathbf{W}_{xk},\mathbf{b}_{hk}\}$.
$g_{kj}(\cdot)$ represents the emission function to link the hidden
variable to the output variable. For example, $g_{kj}(\cdot)$ can
be a linear function with model parameters $\boldsymbol{\theta}_{k}^{g}=\{\mathbf{V}_{yk},\mathbf{b}_{gk}\}$
or nonlinear functions such as neural networks. Here, we denote $\boldsymbol{\theta}_{k}=\{\boldsymbol{\theta}_{k}^{h},\boldsymbol{\theta}_{k}^{g}\}$
as the model parameters for stage $k$. Furthermore, if we use one-layer
neural networks for both transition and emission, the model does share
some similarity with RNN models. However, the major limitation of
using RNN in MMS is that different manufacturing stages are inherently
different. The underlying physics is entirely different for each stage
which not only results in the different transition parameters $\boldsymbol{\theta}_{k}^{h}=\{\mathbf{U}_{hk},\mathbf{W}_{xk},\mathbf{b}_{hk}\}$
and emission parameters $\boldsymbol{\theta}_{k}^{g}=\{\mathbf{V}_{yk},\mathbf{b}_{gk}\}$.
RNN also assumes the same set of variables are predicted in each time.
However, in MMS, different quality inspection sensors are set up in
each manufacturing stage, denoted by $\mathbf{y}_{k}$. Finally, RNN
is a complicated model and can not achieve input and output variable
selection as the proposed approach. More discussion of the relationship
and differences of the one-layer version of the proposed DMMTL and
RNN are shown in Appendix \ref{subsec: RNN}. 

\begin{figure}
\centering\includegraphics[width=1\linewidth]{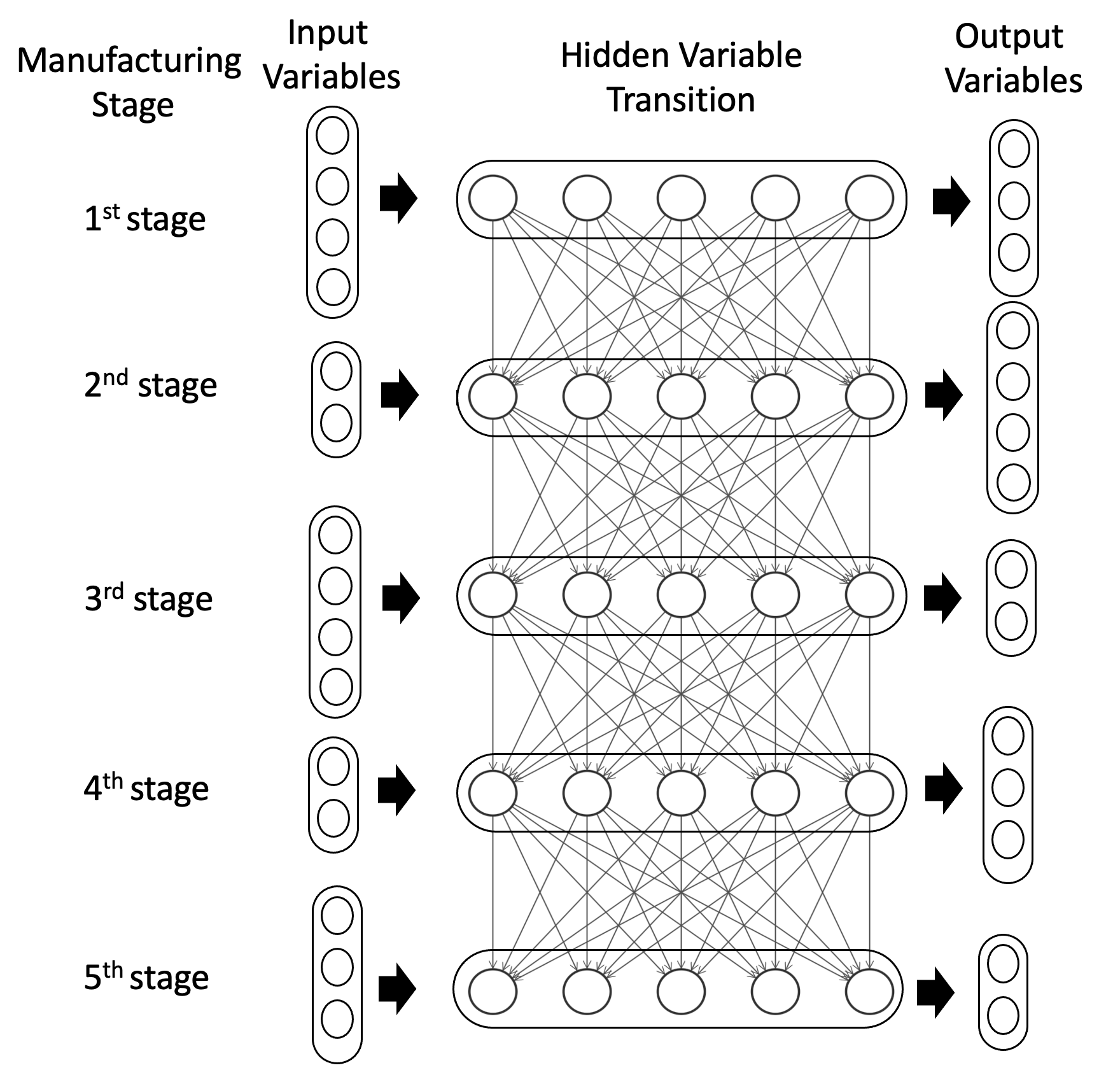}

\caption{Deep Neural Network Structure for the Proposed Method}

\label{Fig: DNN}
\end{figure}

The benefit of the proposed method is also its ultimate flexibility
of plugging in any differentiable functions as $f_{k}(\cdot)$ and
$g_{k}(\cdot)$. For example, depending on different applications,
we can either use simpler models (e.g., linear models) or more complicated
models (e.g., deep neural networks). As an example, two-layer transition
and emission networks are shown as follows:
\begin{align}
\mathbf{h}_{k} & =\sigma(\mathbf{U}_{hk}^{2}\mathbf{h}_{k}^{1}+\mathbf{b}_{hk}^{2}),\mathbf{h}_{k}^{1}=\sigma(\mathbf{W}_{xk}\mathbf{x}_{k}+\mathbf{U}_{hk}^{1}\mathbf{h}_{k-1}^{2}+\mathbf{b}_{hk}^{1})\nonumber \\
\mathbf{y}_{k} & =\mathbf{V}_{yk}^{2}\mathbf{y}_{k}^{1}+\mathbf{b}_{gk}^{2}+\epsilon_{yk},\mathbf{y}_{k}^{1}=\sigma(\mathbf{V}_{yk}^{1}\mathbf{h}_{k}+\mathbf{b}_{gk}^{1})\label{eq: Twolayer}
\end{align}
In this case, the stage transition model parameters are $\boldsymbol{\theta}_{k}^{h}=\{\mathbf{U}_{hk}^{1},\mathbf{U}_{hk}^{2},\mathbf{W}_{xk},\mathbf{b}_{hk}^{1},\mathbf{b}_{hk}^{2}\}$,
and the emission parameters $\boldsymbol{\theta}_{k}^{g}=\{\mathbf{V}_{yk}^{1},\mathbf{V}_{yk}^{2},\mathbf{b}_{gk}^{1},\mathbf{b}_{gk}^{2}\}$
and $\boldsymbol{\theta}_{k}=\{\boldsymbol{\theta}_{k}^{h},\boldsymbol{\theta}_{k}^{g}\}$.
In general, we can use a neural network with depth $D_{1}$ to model
for the transition function and a neural network with depth $D_{2}$
to for the emission function. In this case, the model parameter is
$\boldsymbol{\theta}_{k}=\{\mathbf{W}_{xk},\{\mathbf{U}_{hk}^{d},\mathbf{b}_{hk}^{d}\}_{d=1,\cdots,D_{1}},\{\mathbf{V}_{ykj}^{d},\mathbf{b}_{gk}^{d}\}_{d=1,\cdots D_{2}}\}$.
To show the relationship of the proposed framework and the deep neural
network, we also visualized the architecture of the proposed methods
in Figure \ref{Fig: DNN}. In this figure, we showed a special case
where only a single layer neural network is used for the transition
between the hidden variables. In this case, the number of transition
layers is exactly the number of manufacturing stages. There are some
additional layers models the relationship of the input variables/output
variables with the hidden variables. 

In the proposed DMMTL, instead of modeling each $P(y_{kj}|\{\mathbf{x}_{1},\cdots,\mathbf{x}_{k}\})$
individually, we assume that the hidden state $\mathbf{h}_{k}$ is
learned through the model such that it compresses all the necessary
information to predict the current stage output $\mathbf{y}_{k}$
and future stage output $\mathbf{y}_{k'}$ for $k'>k$. There are
two benefits of using the latent variable $\mathbf{h}_{k}$ rather
than the original data $\{\mathbf{x}_{1},\cdots,\mathbf{x}_{k}\}$:
1) \textit{Dimension reduction} in the sequential transition model:
$\{\mathbf{x}_{1},\cdots,\mathbf{x}_{k}\}$ is typically very high-dimension,
especially for the later stages when $k$ is large. Here the model
creates the low-dimensional hidden state variable $\mathbf{h}_{k}$
to compress all the necessary information from $\{\mathbf{x}_{1},\cdots,\mathbf{x}_{k}\}$.
Therefore, the conditional probability can be compactly represented
by $P(y_{kj}|\{\mathbf{x}_{1},\cdots,\mathbf{x}_{k}\})=P(y_{kj}|\mathbf{h}_{k})$.
2) \textit{Shared representation} for multi-task learning: Here $\mathbf{h}_{k}$
itself is used to predict all the output variables $y_{kj}$ in stage
$k$, which is especially helpful when the output variables $y_{kj}$
in each stage $k$ are correlated. By assuming that different output
variables are conditionally independent given the hidden state variables
$\mathbf{h}_{k}$, the architecture of the model is shown in \Cref{Fig:
Architecture}.

\begin{figure}
\centering \begin{tikzpicture}
\tikzstyle{main}=[circle, minimum size = 10mm, thick, draw =black!80, node distance = 8.5mm]
\tikzstyle{connect}=[-latex, thick]
\tikzstyle{box}=[rectangle, draw=black!100]

  \node[main,fill=black!10] (x1) {$\mathbf{x_1}$};
  \node[main,fill=black!10] (x2) [right=of x1] {$\mathbf{x_2}$};
  \node[main,fill=black!10] (x3) [right=of x2] {$\mathbf{x_3}$};
  \node[main,fill=black!10] (xt) [right=of x3] {$\mathbf{x_k}$};
  
  \node[main] (h1) [below=of x1] {$\mathbf{h_1}$};
  \node[main] (h2) [below=of x2] {$\mathbf{h_2}$};
  \node[main] (h3) [below=of x3] {$\mathbf{h_3}$};
  \node[main] (ht) [below=of xt] {$\mathbf{h_k}$};
  
  \node[main,fill=black!10] (y1) [below=of h1] {$\mathbf{y_1}$};
  \node[main,fill=black!10] (y2) [below=of h2] {$\mathbf{y_2}$};
  \node[main,fill=black!10] (y3) [below=of h3] {$\mathbf{y_3}$};
  \node[main,fill=black!10] (yt) [below=of ht] {$\mathbf{y_k}$};
  
  \path (h3) -- node[auto=false]{\ldots} (ht);
  \path (x1) edge [connect] (h1);
  \path (h1) edge [connect] (y1);
  \path (x2) edge [connect] (h2);
  \path (h2) edge [connect] (y2);
  \path (x3) edge [connect] (h3);
  \path (h3) edge [connect] (y3);
  \path (xt) edge [connect] (ht);
  \path (ht) edge [connect] (yt);
  \path (h1) edge [connect] (h2);
  \path (h2) edge [connect] (h3);
  \plate {pk} {(xt)(yt)(ht)} {};
  
  \node[main,fill=black!10] (xk1) [right= of xt]{${x_{k1}}$};
  \node[main,fill=black!10] (xk2) [right=of xk1]{${x_{k2}}$};
  \node[main,fill=black!10] (xkn) [right=of xk2] {${x_{kn_k}}$};
  \path (xk2) -- node[auto=false]{\ldots} (xkn);

  \node[main] (hk) [below=of xk2] {${h_k}$};
  \path (xk1) edge [connect] (hk);
  \path (xk2) edge [connect] (hk);
  \path (xkn) edge [connect] (hk);
  
  \node[main,fill=black!10] (yk1) [below=of xk1,yshift = -20mm]{${y_{k1}}$};
  \node[main,fill=black!10] (yk2) [below=of xk2,yshift = -20mm]{${y_{k2}}$};
  \node[main,fill=black!10] (ykn) [below=of xkn,yshift = -19mm] {${y_{kn_k}}$};
  \path (yk2) -- node[auto=false]{\ldots} (ykn);
  \path (hk) edge [connect] (yk1);
  \path (hk) edge [connect] (yk2);
  \path (hk) edge [connect] (ykn);
  
  \plate {pkfull} {(xk1)(xk2)(xkn)(hk)(yk1)(yk2)(ykn)} {Stage k};
  
  \draw[-,dashed] (6.3,0.5) -- (6.9,0.6);
  \draw[-,dashed] (6.2,-4.6) -- (6.9,-5.1);

\end{tikzpicture} \caption{Architecture of the Proposed DMMTL}
\label{Fig: Architecture}
\end{figure}
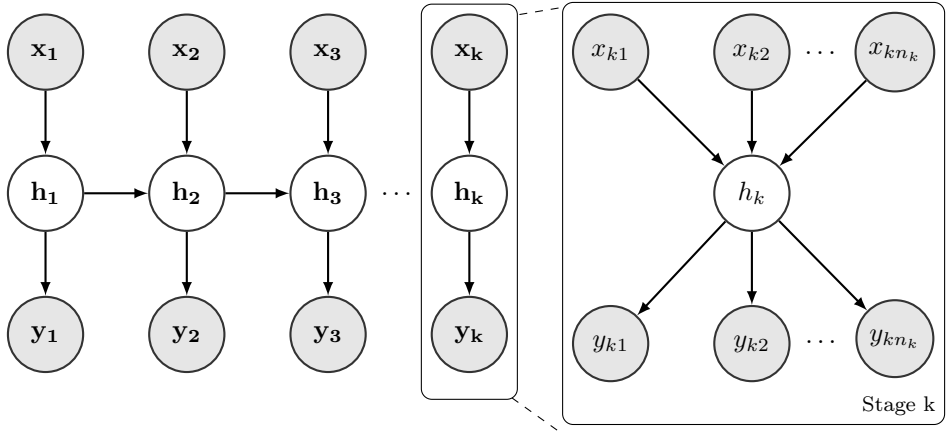

The benefit of introducing this recursive structure and the hidden
state representation $\mathbf{h}_{k}$ is that the negative joint
log likelihood $\mathcal{L}(\boldsymbol{\Theta};\mathcal{X},\mathcal{Y})$
can be decomposed in each stage $k$ and each sensor $j$ as 
\begin{equation}
\mathcal{L}(\mathbf{\Theta};\mathcal{X},\mathcal{Y})=-\sum_{k=1}^{K}\log P(\mathbf{y}_{k}|\mathbf{h}_{k};\boldsymbol{\Theta})=-\sum_{k=1}^{K}\sum_{j=1}^{n_{y,k}}\log P(y_{kj}|\mathbf{h}_{k};\boldsymbol{\Theta})\propto\sum_{k=1}^{K}\sum_{j=1}^{n_{y,k}}L_{k}(\mathbf{e}_{k};\Theta),\label{eq: stagedecomp}
\end{equation}
where $e_{kj}=y_{kj}-g_{kj}(\mathbf{h}_{k};\boldsymbol{\Theta})$
and $L_{k}(\cdot)$ is the negative log-likelihood of the noise distribution.
For example, for Gaussian noise, we can set $L_{k}(\mathbf{e}_{k};\Theta)=\|\mathbf{e}_{k}\|^{2}$.
We will discuss in more detail how to define $L_{k}(\mathbf{e}_{k};\Theta)$
in MMS. Finally, to make the model interpretable and to prevent overfitting,
we propose to minimize the following loss function: 
\begin{equation}
\min_{\Theta}\mathcal{L}(\Theta)+\mathcal{R}(\Theta),\label{eq: fullMSM}
\end{equation}
where $\Theta=\{\boldsymbol{\theta}_{1},\cdots,\boldsymbol{\theta}_{K}\}$.
$\mathcal{L}(\Theta)=\sum_{k=1}^{K}L_{k}(\boldsymbol{\theta}_{k})$
is the likelihood loss function, $\mathcal{R}(\Theta)=\sum_{k=1}^{K}R_{k}(\boldsymbol{\theta}_{k})$
is the regularization function. Here, $R_{k}(\boldsymbol{\theta}_{k})$
and $L_{k}(\boldsymbol{\theta}_{k})$ are defined as the regularization
term and the loss function in stage $k$, which will be defined in
more detail later. \Cref{eq: fullMSM} can also be seen as a multi-layer
neural network, where the architecture of the neural network structure
depends on the physical layout of the manufacturing system. For example,
each layer (or a set of layers) represents one stage of the manufacturing
system, with the emission network output the prediction of output
sensing variables (i.e., quality indices measured at each stage),
and the transmission network passes the information to the next stage.
We propose to combine the process variables $\mathbf{x}_{k}$ or quality
variables $\mathbf{y}_{k}$ for all $K$ stages in a multi-task learning
framework in order to optimize $\boldsymbol{\theta}_{k}$ in an end-to-end
fashion. 

It is also interesting to compare the proposed DMMTL in \Cref{eq:
fullMSM} to the independent modeling approach, where each output
variable $y_{kj}$ at stage $k$ and output $j$ is modeled independently.
Because such an independent modeling approach needs to introduce new
model parameters $\boldsymbol{\theta}_{kj}$ for each output variable
$j$ in each stage $k$, for a $K$-stage system with $n_{y}$ output
sensors and $n_{x}$ input sensors in each stage, it requires to have
$O(K^{2}n_{x}n_{y})$ model parameters in total. These model parameters
are normally time-consuming to train and can lead to the over-fitting
problem. On the other hand, since our proposed framework reduces the
hidden state dimension to $n_{h}$ using $\boldsymbol{\theta}_{k}^{h},k=1,\cdots,K$
with $O(Kn_{h}(n_{x}+n_{y}))$ number of variables, which is typically
much smaller than $O(K^{2}n_{x}n_{y})$ given $n_{h}\ll\min(Kn_{x},Kn_{y})$.
Therefore, a shared representation yields a more compact representation
with better memory efficiency. 

Furthermore, we would like to propose the loss function $L_{k}(\cdot)$
and regularization $R_{k}(\boldsymbol{\theta}_{k})$ in \Cref{eq:
generalrecursive} that leads to a better engineering interpretation.
In addition, we use the regularized model to select key input variables
and output variables. More specifically, we propose the group lasso
penalty \citep{yuan2006model} and robust statistics for a more interpretable
transition model, allowing us to perform input and output variable
selection. 

\paragraph{Group Sparsity Penalty }

In MMS, many input variables are irrelevant in relation to predicting
output variables. Therefore, these input variables should not affect
the hidden states. To select the important sensors in each stage,
we propose to use the $L_{2,1}$ nom on $\mathbf{W}_{xk}^{T}$ to
encourage the model to only select the most important sensors. In
the literature, the $L_{2,1}$ norm is defined as $\|\mathbf{W}\|_{2,1}=\sum_{i}\sqrt{\sum_{j}W_{ij}^{2}}$,
which penalizes the entire row of the matrix $\mathbf{W}$ to be zero.
Here, we propose to add this $L_{2,1}$ norm on the transposed coefficient
$\mathbf{W}_{xk}$ on (\ref{eq: stagedecomp}), which penalizes the
entire column of matrix $\mathbf{W}_{xk}$ to be zero. In other words,
if we define the $i^{th}$ column of $\mathbf{W}_{xk}$ is $\mathbf{w}_{i,xk}$,
$\|\mathbf{W}_{xk}^{T}\|_{2,1}=\sum_{j}\|\mathbf{w}_{i,xk}\|$. 

This penalty not only controls the model flexibility but also can
lead to a more interpretable result due to the sensor selection power
as follows:

\begin{equation}
R_{k}(\boldsymbol{\theta}_{k})=\lambda_{x}\|\mathbf{W}_{xk}^{T}\|_{2,1}+\frac{\lambda}{2}\|\Theta\|^{2},\label{eq: Regularized}
\end{equation}
where $\|\cdot\|_{2}$ is the $L_{2}$ norm and $\|\mathbf{W}_{xk}^{T}\|_{2,1}$
tends to penalize the entire column of $\mathbf{W}_{xk}$ to be $0$.
For example, if the $i^{th}$ column of $\mathbf{W}_{xk}$ is $0$,
the $i^{th}$ input variable of $x_{ki}$ in stage $k$ is not selected,
which means this input variable $x_{ki}$ is not important to the
prediction of any output variables in the following stages. Furthermore,
$\frac{\lambda}{2}\|\Theta\|^{2}$ is added for the general $L_{2}$
penalty to prevent overfitting. We will discuss how to identify the
most important input variable for each individual output variable
in \Cref{sec: EstimationDiagnosis}. These regularization terms
enforce that only some of the input variables or hidden variables
are used in the prediction, which models the weakly correlated patterns
and improves the interpretability of the model. Finally, we will use
the validation prediction error to select the best tuning parameters
$\lambda_{x}$, $\lambda$ and $\gamma$. 

\paragraph{Robust Regression }

The most commonly defined loss function is the sum of square error
(SSE) loss defined as $L_{k}(\mathbf{e})=\|\mathbf{e}\|^{2}$. However,
the proposed DMMTL is a multi-task learning framework that focuses
on predicting multiple tasks (i.e., quality variables) simultaneously.
In reality, due to the lack of sensing powers, many output variables
simply cannot be predicted by the input variables no matter what models
we use. Using these non-informative quality variables may not help
or even corrupt the training results. Here, the goal is to derive
the loss function such that the model is robust to these unrelated
tasks or achieve a better balance between tasks. Therefore, we will
compare the use of the Huber loss function $L_{k}(e)=\rho(e)$ as
defined in \Cref{eq: huber} with the traditional sum of square
loss function. 
\begin{equation}
\rho(e)=\begin{cases}
\|e\|^{2} & |e|\leq\frac{\gamma}{2}\\
\gamma|e|-\frac{\gamma^{2}}{4} & |e|>\frac{\gamma}{2}
\end{cases}.\label{eq: huber}
\end{equation}
Huber loss can be used instead of the mean-squared error. The Huber
loss function uses a linear function when the difference is large
which enables a more robust estimation. Furthermore, we find that
it can also help the model identify and focus more directly on the
related output variables by being more robust to the unrelated output
variables. We will discuss how to optimize the model parameters efficiently
in \Cref{subsec:Optimization-Algorithm}. 

\subsection{Optimization Algorithm \label{subsec:Optimization-Algorithm} }

It is worth noting that problem (\ref{eq: fullMSM}) has a non-smooth
loss function penalty such as $\lambda_{x}\|\mathbf{W}_{xk}^{T}\|_{2,1}$.
In literature, the stochastic sub-gradient algorithm can be used to
handle the non-smooth penalty. However, the convergence speed of the
stochastic sub-gradient algorithm is typically slow. Therefore, to
address the non-smooth loss function, we propose to combine the stochastic
proximal gradient algorithm and block coordinate algorithm for efficient
optimization. 

To efficiently optimize the problem, we will discuss the use of the
stochastic proximal gradient as follows. First, we will first establish
the equivalency of the proposed algorithm by introducing another set
of outlier variables $\mathcal{A}=\{a_{k,j}\}_{k,j}$. Here, $a_{k,j}$
represents the outlier in stage $k$ and sensor $j$. If $a_{k,j}=0$,
it implies that there is no outlier in stage $k$ and sensor $j$
for the current sample. However, if $a_{k,j}\neq0$, it implies that
the outlier occurs for this sample. 
\begin{prop}
Solving $\boldsymbol{\Theta}$ in (\ref{eq: fullMSM}) will give the
same solution as solving $\boldsymbol{\Theta}$ in (\ref{eq: Opt}).
\begin{equation}
\min_{\boldsymbol{\Theta},\{a_{kj}\}}\mathcal{L}(\Theta,\mathcal{A})+\lambda_{x}\|\mathbf{W}_{xk}^{T}\|_{2,1}+\frac{\lambda}{2}\|\Theta\|^{2}+\gamma\sum_{k,j}\|a_{k,j}\|_{1},\label{eq: Opt}
\end{equation}
where $\mathcal{L}(\Theta,\mathcal{A})=\sum_{k,j}\|y_{kj}-g_{kj}(\mathbf{h}_{k};\boldsymbol{\Theta})-a_{kj}\|^{2}$
is the loss function, $\mathcal{A}=\{a_{k,j}\}_{k,j}$. 
\end{prop}

It worth noting that both $y_{kj}$ and $a_{kj}$ are scalar if only
a single sample is used. However, if multiple samples are used, $y_{kj}$
and $a_{kj}$ are vectors. Please see Appendix D for more details
to generalize the proposed methods to the mini-batch version. 

The proof relies on the equivalency of the Huber loss function and
the sparse outlier decomposition and has been proved in \citep{mateos2011robust}.
The benefit of using (\ref{eq: Opt}) is that $\mathcal{R}(\Theta,\mathcal{A})$
is continuous and differentiable so that the back-propagation algorithm
can be used efficiently. We will show how to handle non-differentiable
components of $\sum_{k,j}\|a_{kj}\|_{1}$ and $\|\mathbf{W}_{xk}^{T}\|_{2,1}$
by using Block Coordinate algorithm to update each set of variables
$\mathcal{A}$, $\mathbf{W}_{xk}$, and $\Theta$ (excluding $\mathbf{W}_{xk}$)
iteratively until convergence. Proposition 2 shows how to solve $\mathcal{A}$
and $\mathbf{W}_{xk}$, given the other variables are fixed. 
\begin{prop}
In the $t^{th}$ iteration, solving \textup{$a_{kj}$ given $\Theta=\Theta^{(t)}$
in (\ref{eq: Opt}) can be derived analytically as}
\[
a_{kj}^{(t)}=S_{\gamma/2}(y_{kj}-g_{kj}(\mathbf{h}_{k};\boldsymbol{\Theta}^{(t)})).
\]
\textup{Given $\Theta=\Theta^{(t)}$ and} $\mathcal{A}^{(t)}$, the
upper bound of (\ref{eq: Opt}), defined as \textup{
\begin{align*}
\min_{\{\mathbf{w}_{i,xk}\}}\mathcal{L}(\Theta^{(t-1)},\mathcal{A}^{(t)})+\sum_{i,k}\frac{\partial\mathcal{L}(\Theta^{(t-1)},\mathcal{A}^{(t)})}{\partial\mathbf{w}_{i,xk}}\left(\mathbf{w}_{i,xk}-\mathbf{w}_{i,xk}^{(t-1)}\right)\\
+\frac{L}{2}\sum_{i,k}\|\mathbf{w}_{i,xk}-\mathbf{w}_{i,xk}^{(t-1)}\|^{2}+\lambda_{x}\sum_{i,k}\|\mathbf{w}_{i,xk}\|_{2}+\frac{\lambda}{2}\sum_{i,k}\|\mathbf{w}_{i,xk}\|^{2}.
\end{align*}
 in the proximal gradient algorithm, can be solved by }
\end{prop}

\[
\mathbf{w}_{i,xk}^{(t)}=S_{\frac{\lambda_{x}}{L+\lambda}}(\frac{L}{L+\lambda}(\mathbf{w}_{i,xk}^{(t-1)}-\frac{1}{L}\frac{\partial\mathcal{L}(\Theta^{(t)},\mathcal{A}^{(t)})}{\partial\mathbf{w}_{i,xk}})).
\]
Here, $S_{\gamma}(x)=\mathrm{sgn}(x)(\left|x\right|-\gamma)_{+}$
is the soft thresholding operator, in which $\mathrm{sgn}(x)$ is
the sign function and $x_{+}=\max(x,0)$. $\Theta^{(t)}$, $\mathcal{A}^{(t)}$,
and $\mathbf{w}_{i,xk}^{(t)}$ are the corresponding values of the
model parameters in the $t^{th}$ iterations. $L$ is the Lipschitz
constance of the function $\mathcal{L}(\cdot)$. 

The proof is given in Appendix \ref{sec:proximalGrad}. The gradient
$\frac{\partial\mathcal{L}(\Theta^{(t)},\mathcal{A}^{(t)})}{\partial\mathbf{w}_{i,xk}}$
can be computed from the back-propagation algorithm, which is detailed
in Appendix \ref{sec:Gradient}. Finally, since the loss function
is differential for the parameter $\tilde{\Theta}$, defined as $\boldsymbol{\Theta}$
excluding $\mathcal{W}=\{W_{i,xk}\}_{i,k}$, the standard stochastic
gradient algorithm can be applied given $\mathcal{W}=\mathcal{W}^{(t)}$
and $\mathcal{A}=\mathcal{A}^{(t)}$ as follows:
\[
\tilde{\Theta}=\tilde{\Theta}-c\frac{\partial\mathcal{L}(\Theta,\mathcal{W}^{(t)},\mathcal{A}^{(t)})}{\partial\tilde{\Theta}},
\]
where $c$ is the step length. Finally, the mini-batch version of
the algorithm can also be derived with a subset of samples $\{\mathcal{X}^{n},\mathcal{Y}^{n}\}_{n\in\mathcal{N}^{t}}$
in iteration $t$. More details of using stochastic optimization algorithm
is also shown in Appendix \ref{sec:Gradient}. 

The algorithm is summarized in \Cref{algo:DMMTL}. Given the non-convex
formulation of deep neural networks, there is no guarantee that the
algorithm would converge to the the global optimum. However, we find
out that optimization algorithms typically perform reasonably well.
In case the training failed to converge to a good optimum (measured
by the validation accuracy), the training can be restarted with a
new random initialization point. 

\begin{algorithm}
\DontPrintSemicolon 
\KwIn{ $\mathcal{X}=\{\mathbf{x}_{1},\cdots,\mathbf{x}_{K}\}$ are the input/process sensing variables for $K$ stages }
\KwOut{$\mathcal{Y}=\{\mathbf{y}_{1},\cdots,\mathbf{y}_{K}\}$ are the ouput/quality sensing variables for $K$ stages }
\For{$t = 1,2,\cdots$} {
Compute $\frac{\partial\mathcal{L}(\Theta,\mathcal{W}^{(t)},\mathcal{A}^{(t)})}{\partial {\Theta}}$ based on back-propagation through a random subset of samples $\{\mathcal{X}^{n},\mathcal{Y}^{n}\}_{n\in\mathcal{N}^{t}}$. \\
\For{$k = 1,2,\cdots K$} {
\For{$j=1,\cdots, N_{y,k}$}{
Update $a_{kj}^{(t)}=S_{\gamma/2}(y_{kj}-g_{kj}(\mathbf{h}_{k};\boldsymbol{\Theta}^{(t)})).$ \\
}
\For{$i=1,\cdots,N_{x,k}$}{
Update $\mathbf{w}_{i,xk}^{(t)}=S_{\frac{\lambda_{x}}{L+\lambda}}(\frac{L}{L+\lambda}(\mathbf{w}_{i,xk}^{(t-1)}-\frac{1}{L}\frac{\partial\mathcal{L}(\Theta^{(t)},\mathcal{A}^{(t)})}{\partial\mathbf{w}_{i,xk}})).$
}
}
Update $\tilde{\Theta}=\tilde{\Theta}-c\frac{\partial\mathcal{L}({\Theta},\mathcal{W}^{(t)},\mathcal{A}^{(t)})}{\partial\tilde{\Theta}}$ 
}
\caption{Optimization Algorithm for DMMTL}
\label{algo:DMMTL}
\end{algorithm}

\subsection{Tuning Parameter Selection}

In this section, we would like to discuss the selection of tuning
parameters. Overall, we need to decide the following tuning parameters:
1) Number of dimensions of the hidden vector $\mathbf{h}_{k}$. For
the dimensionality of the hidden vector $\mathbf{h}_{k}$, in principle,
we can vary the number of neurons for $\mathbf{h}_{k}$ for different
stage $k$. Here, we would like to mention that in this paper, we
use the same dimensionality of the hidden vector $\mathbf{h}_{k}$
for different stage $k$ for simplicity. However, we will use the
regularization term to control the amount of information that flows
into the network. 2) Tuning parameter $\lambda_{x}$ and $\lambda$.
Here, the $\lambda_{x}$ is used to control the sparsity of the input
variables. For example, increasing $\lambda_{x}$ will lead to a more
sparse selection of the input variables at different stages. $\lambda$
is used to regularize the $L_{2}$ norm of all the parameters. 3)
The depth of neural network architectures for each stage. Again, in
principle, we can use different layer depths for different manufacturing
stages. For example, if one particular manufacturing stage is more
complex, we can actually increase the number of layers for such a
stage. In our simulation study and case study, we find that a one-layer
neural network for each stage is enough to cover most of the cases. 

Finally, to select all these parameters, we propose to use a set of
validation dataset $\{\mathcal{X}^{val},\mathcal{Y}^{val}\}$. Furthermore,
we will use a randomized search of the tuning parameter space with
the prediction accuracy of the validation set $\|\mathcal{Y}^{val}-\mathcal{\hat{Y}}^{val}\|^{2}$
as the metric to select the best combinations of the tuning parameters. 

\section{Improve Model Local and Global Interpretability \label{sec: EstimationDiagnosis}}

After the model is derived in \Cref{eq: fullMSM} and the transition
and emission are defined in Equations (\ref{eq: Regularized}) and
(\ref{eq: huber}), we will discuss how to improve the model interpretability
by developing novel techniques for input variables identifications.
Moreover, we aim to develop an interpretation module in this chapter
to understand what happens exactly in the black-box model. In literature,
there are two types of interpretability, global interpretability and
local interpretability. Global interpretability refers to understand
how the model makes decisions, based on a holistic view of its features
and each of the learned components such as weights, other parameters,
and structures, as defined in Chapter 2.3.2 in \citep{molnar2020interpretable}.
In the context of MMS, global interpretability refers to identify
important input variables that are important for any output variables
for all samples. We will discuss how to achieve global interpretability
in the proposed DMMTL method in Section \ref{subsec:Input-Variable-Identification}. 

Local interpretability refers to understand how the model makes decisions
based on each individual sample. Local interpretability is quite important,
given that the different output variables for different samples may
depend on different sets of input variables, as defined in Chapter
2.3.4 in \citep{molnar2020interpretable}. In the context of MMS,
local interpretability refers to identify important input variables
(e.g., process variables) for each output variable (e.g., quality
index) for each individual sample (i.e., or a subset of samples).
We will discuss how to achieve the local interpretability in the proposed
DMMTL method in Section \ref{subsec:Diagnostics-According-to}. 

\subsection{Improve Model Local Interpretability \label{subsec:Input-Variable-Identification}}

In this subsection, we will discuss how to achieve global model interpretability
by identifying the input variables contributing to the prediction
of the entire MMS systems by examining the model coefficients. In
general, the $L_{2,1}$ penalty is able to pick up the most important
input variables for each stage automatically and the non-zero value
of the $L_{2}$ norm of different columns of $\mathbf{W}_{xk}$, or
namely $\|\mathbf{w}_{i,xk}\|_{2}$, correspond to the most important
input variables in stage $k$. These important input variables are
selected because they contribute significantly to the entire MMS systems\textbf{
(}i.e., any output variable in the future stages). We will then discuss
how to achieve local interpretation by identifying important sensing
variables with respect to each individual output sensor (i.e., diagnostics)
for a single sample.

\subsection{Improve Model Global Interpretability \label{subsec:Diagnostics-According-to}}

In this subsection, we will discuss how to achieve local model interpretability
by identifying the important input variables that relate the most
to one specific quality index for any particular subset of samples.
Motivated by \citep{apley2016visualizing}, we propose the gradient
tracking technique to achieve the input variable selection for a selected
output variable for a particular sample for local interpretability.
This is done through tracking back the gradient of the identified
output variable $y_{kj}$ according to each individual input variable
by back-propagation through the output linkage function $y_{kj}=g_{kj}(\mathbf{h}_{k};\mathbf{\theta}_{k}^{g})$
and the state transition matrix $\mathbf{h}_{k}=f_{k}(\mathbf{h}_{k-1},\mathbf{x}_{k},\mathbf{y}_{k-1};\mathbf{\theta}_{k}^{h})$.
If we choose a linear function for both $g_{kj}(\cdot)$ and $f_{k}(\cdot)$,
the relationship between $y_{kj}$ and each individual $x_{k}$ is
also linear. However, for a nonlinear output function and state transition
matrix, the exact functional form of each $y_{kj}$ and input variable
$\mathbf{x}$ can be quite complex. To analyze the relationship between
the input and output variables, we propose to use the Taylor expansion
of $y_{kj}$ according to the input variable $\mathbf{x}$ as follows:

\begin{equation}
y_{pq}(\mathbf{x}+\Delta\mathbf{x})=\sum_{k=1}^{K}\sum_{i=1}^{n_{xk}}\frac{\partial y_{pq}}{\partial x_{ki}}\Delta x_{ki}+\sum_{k}\sum_{k'}\sum_{i}\sum_{i'}\Delta x_{ki}\frac{\partial^{2}y_{pq}}{\partial x_{ki}\partial x_{k'i'}}\Delta x_{k'i'}+O(\Delta x_{ki}^{3})\label{eq: sensitivity}
\end{equation}
Therefore, the relative importance of a sensor can be computed by
the gradient $\frac{\partial y_{pq}}{\partial x_{ki}}$. The first-order
gradient information can be computed through the back-propagation
through the sequential stages, as 
\begin{equation}
\frac{\partial y_{pq}}{\partial\mathbf{x}_{k}}=\frac{\partial y_{pq}}{\partial\mathbf{h}_{k'}}\frac{\partial\mathbf{h}_{k'}}{\partial\mathbf{h}_{k'-1}}\cdots\frac{\partial\mathbf{h}_{k+1}}{\partial\mathbf{h}_{k}}\frac{\partial\mathbf{h}_{k}}{\partial\mathbf{x}_{k}}.\label{eq: grad_x}
\end{equation}
The detailed derivation for each component is shown in Appendix \ref{sec:Gradient}.
Because the gradient $\frac{\partial y_{pq}}{\partial\mathbf{x}_{k}}$
also depends on the value of $\mathbf{x}_{k}$, to obtain the relative
importance of each input variable, we propose to compute the sum of
squares of the gradient averaging over either the entire samples as
$\frac{1}{N_{n}}\sum_{n=1}^{N_{n}}(\frac{\partial y_{pq}}{\partial\mathbf{x}_{k}^{n}})^{2}$
or a selected number of defective samples for local interpretation.
It is worth noting that in many applications, the important sensors
may not be consistent across different ranges of $\mathbf{x}_{k}$.
In this case, it may be useful to divide the entire region of $\mathbf{x}_{k}$
into different windows and compute this average for each window.

\section{Simulation \label{sec:Simulation} }

In this section, we simulate a multistage manufacturing process with
$9$ stages. For the $k^{th}$ stage there are $n_{x}=90$ input variables,
denoted as $\mathbf{x}_{k}=(x_{k,1},\cdots,x_{k,n_{x}})$ and $n_{y}=6$
output variables, denoted as $\mathbf{y}_{k}=(y_{k,1},\cdots,y_{k,n_{y}})$.
For the input variable we simulate $x_{k,i}\overset{i.i.d}{\sim}N(0,1)$.
We will discuss how to generate the output variables in three different
scenarios. In all simulation scenarios, we assume there is a hidden
state representation $\mathbf{h}_{k}$. We will assume three different
hidden state $\mathbf{h}_{k}$ transition scenarios.

\paragraph{Case 1. One Unified MMS }

In this case, we generate the MMS with linear hidden state transition
model as $\mathbf{h}_{k}=\mathbf{W}_{xk}\mathbf{x}_{k}+\mathbf{U}_{hk}\mathbf{h}_{k-1}+\mathbf{b}_{k}$.
Furthermore, for each output variable $y_{k,i}$, it is generated
from the $\mathbf{y}_{k}=\mathbf{W}_{yk}\mathbf{h}_{k}+\boldsymbol{\epsilon}_{ki}$,
where $\epsilon_{ki}\sim N(0,\sigma^{2})$ and $\sigma=0.5$. Each
element of $\mathbf{W}_{xk}$ is generated from normal distribution
$N(0,\frac{1}{\sqrt{n_{x}}})$. Each element of $\mathbf{b}_{k}$,$\mathbf{U}_{hk}$,
$\mathbf{W}_{yk}$ is generated from $N(0,\frac{1}{\sqrt{n_{h}}})$,
where $n_{h}$ is the dimensionality of $\mathbf{h}_{k}$. Furthermore,
for each stage $k$, we will generate $15$ out of $n_{x}=90$ as
unimportant sensors for each stage by setting the last $15$ rows
of $\mathbf{W}_{xk}$ to be $0$.

\paragraph{Case 2. Three Parallel Sensor Groups in one MMS}

In this case, we assume that the input variable $\mathbf{x}_{k}$
and output variable $\mathbf{y}_{k}$ can be divided into three groups
as $\mathbf{x}_{k}=\{\mathbf{x}_{k}^{(g)},g=1,2,3\}$ and $\mathbf{y}_{k}=\{\mathbf{y}_{k}^{(g)},g=1,2,3\}$.
For each group, we generate $n_{x}^{(g)}=30$ input variables and
$n_{y}^{(g)}=2$ output variables. For each group $g$, the hidden
state $\mathbf{h}_{k}^{(g)}$ follows its own transition and only
relates to the corresponding output variables $\mathbf{y}_{k}^{(g)}$
as $\mathbf{h}_{k}^{(g)}=\mathbf{W}_{xk}^{(g)}\mathbf{x}_{k}^{(g)}+\mathbf{U}_{hk}^{(g)}\mathbf{h}_{k-1}^{(g)}+\mathbf{b}_{k}^{(g)}$
and $\mathbf{y}_{k}^{(g)}=\mathbf{W}_{yk}^{(g)}\mathbf{h}_{k}^{(g)}+\boldsymbol{\epsilon}_{ki}^{(g)}$,
$g=1,2,3$. In this case, each element of $\mathbf{W}_{yk}^{(g)}$,
$\mathbf{b}_{k}^{(g)}$, and $\mathbf{U}_{hk}^{(g)}$ is generated
from the normal distribution $N(0,\frac{1}{\sqrt{n_{h}^{(g)}}})$,
where $n_{h}^{(g)}$ is the dimensionality of the hidden state $\mathbf{h}_{k}^{(g)}$.
Each element of $\mathbf{W}_{yk}^{(g)}$ is generated from $N(0,\frac{1}{\sqrt{n_{x}^{(g)}}})$
and $\epsilon_{ki}\sim N(0,\sigma^{2})$. Furthermore, for each stage
$k$ and group $g$, we generate $5$ unimportant input variables
for each stage by setting the last $5$ rows of $\mathbf{W}_{xk}^{(g)}$
to be $0$.

\paragraph{Case 3. Three Manufacturing Lines in one MMS}

In this case, we assume that the state transition is only related
to the hidden state that is three-stage apart $\mathbf{h}_{k}=\mathbf{W}_{xk}\mathbf{x}_{k}+\mathbf{U}_{hk}\mathbf{h}_{k-3}+\mathbf{b}_{k}$.
In other words, the three groups of output variables in the stage
$k=1,4,7$, $k=2,5,8$, $k=3,6,9$ are correlated within the group
but independent from each other. However, this relationship is assumed
to be unknown. We generate $n_{x}=90$ input variables and $n_{y}=6$
output variables in each stage. Each element of $\mathbf{W}_{xk}$
is generated from normal distribution $N(0,\frac{1}{\sqrt{n_{x}}})$.
Each element of $\mathbf{b}_{k}$,$\mathbf{U}_{hk}$, $\mathbf{W}_{yk}$
is generated from $N(0,\frac{1}{\sqrt{n_{h}}})$, where $n_{h}$ is
the dimensionality of $\mathbf{h}_{k}$. Furthermore, for each stage
$k$, we generate $15$ unimportant input variables for each stage
by setting the last $15$ rows of $\mathbf{W}_{xk}$ to be $0$.

In all cases, we assume that the relationship between different stages
and different sensing variables are not known. The goal is to predict
each output variable $y_{kj}$, given the input variables up to stage
$k$, denoted by ${\mathbf{x}_{1},\cdots,\mathbf{x}_{k}}$ without
relying on the specific relationship between stages. Finally, we divide
the data into training ($\mathbf{x}^{tr}$, $\mathbf{y}^{tr}$) and
testing ($\mathbf{x}^{te}$, $\mathbf{y}^{te}$) and use the relative
mean of squared error (RMSE) $\sum_{k}\sum_{j}\|y_{k,j}^{te}-\hat{y}_{k,j}^{te}\|^{2}/\|y_{k,j}^{te}-\bar{y}_{k,j}^{tr}\|^{2}$
for performance evaluation. If RMSE is smaller than $1$ it means
that the estimator is able to get some signals by beating the naive
predictor as the training mean. A smaller RMSE indicates better performance.
For benchmark methods, we will compare the proposed MMS model framework
with other prediction methods that focus on modeling each single input
variable individually. The modeling methods we are comparing are linear
regression (LR), elastic net (EN), random forest (RF) and multi-layer
perceptron (MLP). We will also include the stream of variation (SoV)
as an oracle method since it assumes the true transition and output
function is known. For the MLP, we build a 2-layer fully connected
neural network to link the input variables up to stage $k$ to each
output variable $y_{kj}$. We also compare with another Multi-task
learning method, namely the Multi-task Elastic Net (MEN), which combines
the MTL to model the multivariate response in each stage $\mathbf{y}_{k}$
and elastic net for variable selection. MEN assumes that the same
variables should be selected for different tasks/sensors in each stage.
For more details about MEN methods, please check the Appendix . To
be fair, we will look at the supervised learning models to predict
the output sensors in the last stage and look at the magnitude of
the model coefficients for each sensor. The results are shown in \Cref{Table:
RMSE}.

\begin{table}
\caption{Prediction RMSE and standard deviation}
\centering %
\begin{tabular}{|c|c|c|c|}
\hline 
\multirow{1}{*}{} & \multicolumn{3}{c|}{RMSE}\tabularnewline
\hline 
 & Case 1 & Case 2 & Case 3\tabularnewline
\hline 
\hline 
DMMTL & \textbf{0.090 (0.037)} & \textbf{0.138 (0.060)} & \textbf{0.134 (0.057)}\tabularnewline
\hline 
MEN & 0.239 (0.134) & 0.192 (0.108) & 0.166 (0.072)\tabularnewline
\hline 
LR & 0.577 (0.632) & 0.666 (1.173) & 1.213 (1.653)\tabularnewline
\hline 
EN & 0.273 (0.162) & 0.150 (0.090) & 0.273 (0.127)\tabularnewline
\hline 
RF & 0.863 (0.079) & 0.768 (0.123) & 0.822 (0.085)\tabularnewline
\hline 
MLP & 0.728 (0.266) & 0.808 (0.330) & 1.006 (0.413)\tabularnewline
\hline 
\end{tabular}

\label{Table: RMSE}
\end{table}

From \Cref{Table: RMSE}, we see that DMMTL is able to obtain the
smallest RMSE in all cases compared to other benchmark methods. This
superiority is due to the following three reasons: 1) Benefit of modeling
the multiple output variables jointly.  MEN models all output variables
in each stage jointly by assuming that these model coefficients share
the same sparsity structure, which can often lead to better modeling
accuracy. This benefit can be seen in Case 3 by comparing the RMSE
of MEN, about 0.166, with the RMSE of EN, about 0.273. However, we
have to mention that in the case when the output variables are not
correlated,  this could lead to a worse result. For example, in Case
2, there are 3 different variable groups in each manufacturing stage
and EN achieves the RMSE of 0.150, which is better than MEN with the
RMSE as 0.192). 2) Benefit of jointly modeling the manufacturing stages.
 MEN models do not model the output variables in all stages jointly
like DMMTL. In all cases, DMMTL achieves a smaller RMSE than that
of MEN. For example, in Case 1, DMMTL achieves an RMSE of 0.090, which
is smaller than the second best, namely the MEN, which achieves RMSE
of $0.239$. This shows the advantage of modeling all output variables
jointly in all stages. 3) Benefit of sparsity penalty for feature
selection. This benefit can be seen clearly by comparing the EN with
LR and MLP. Without any sparsity constraint, both MLP and LR perform
badly with a much larger variance due to the model over-fitting. In
comparison, EN clearly outperforms these two methods in all cases.

\begin{figure}
\subfloat[Case 1]{\includegraphics[width=0.33\linewidth]{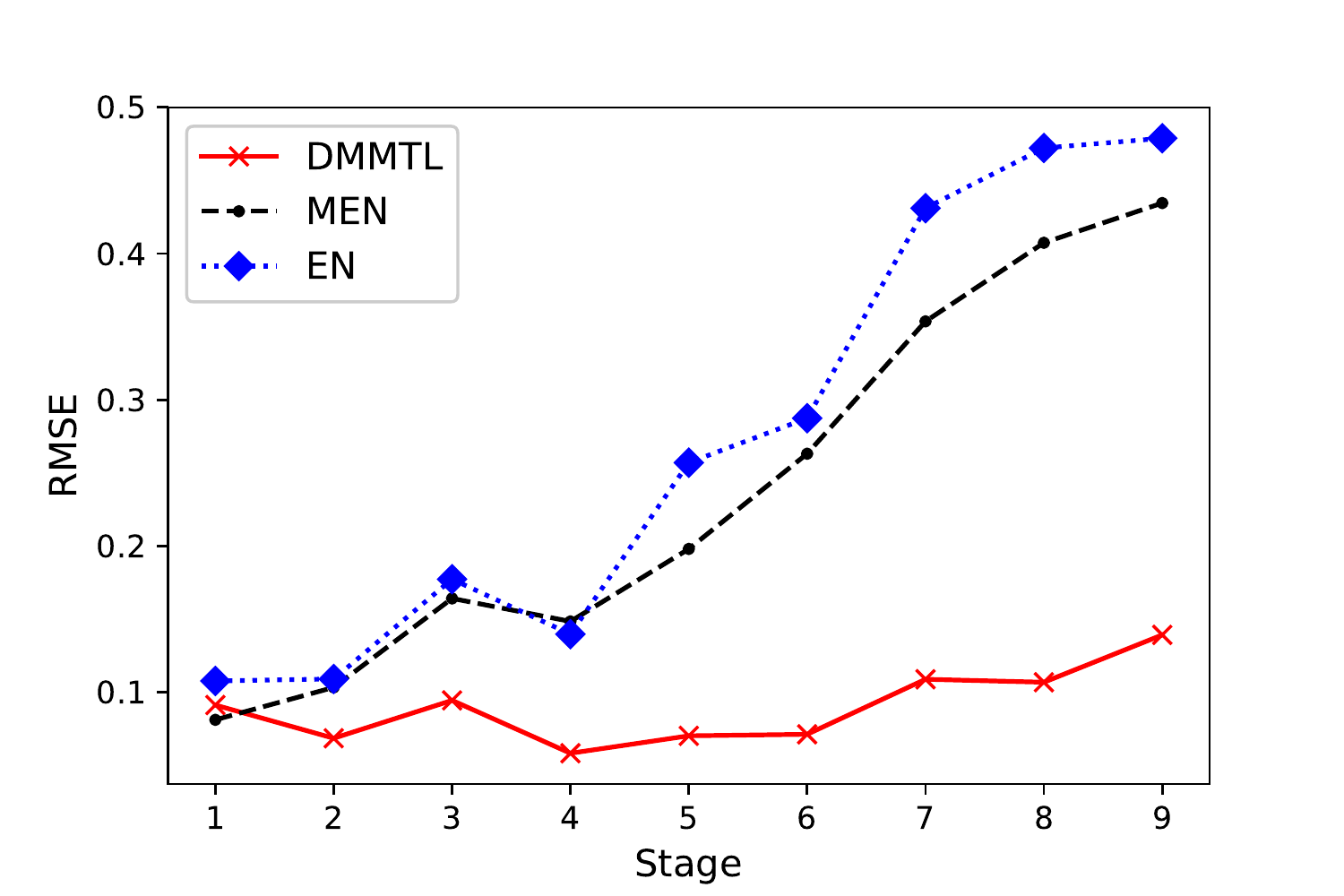}

\label{Fig: RMSEcase1}

}\subfloat[Case 2]{\includegraphics[width=0.33\linewidth]{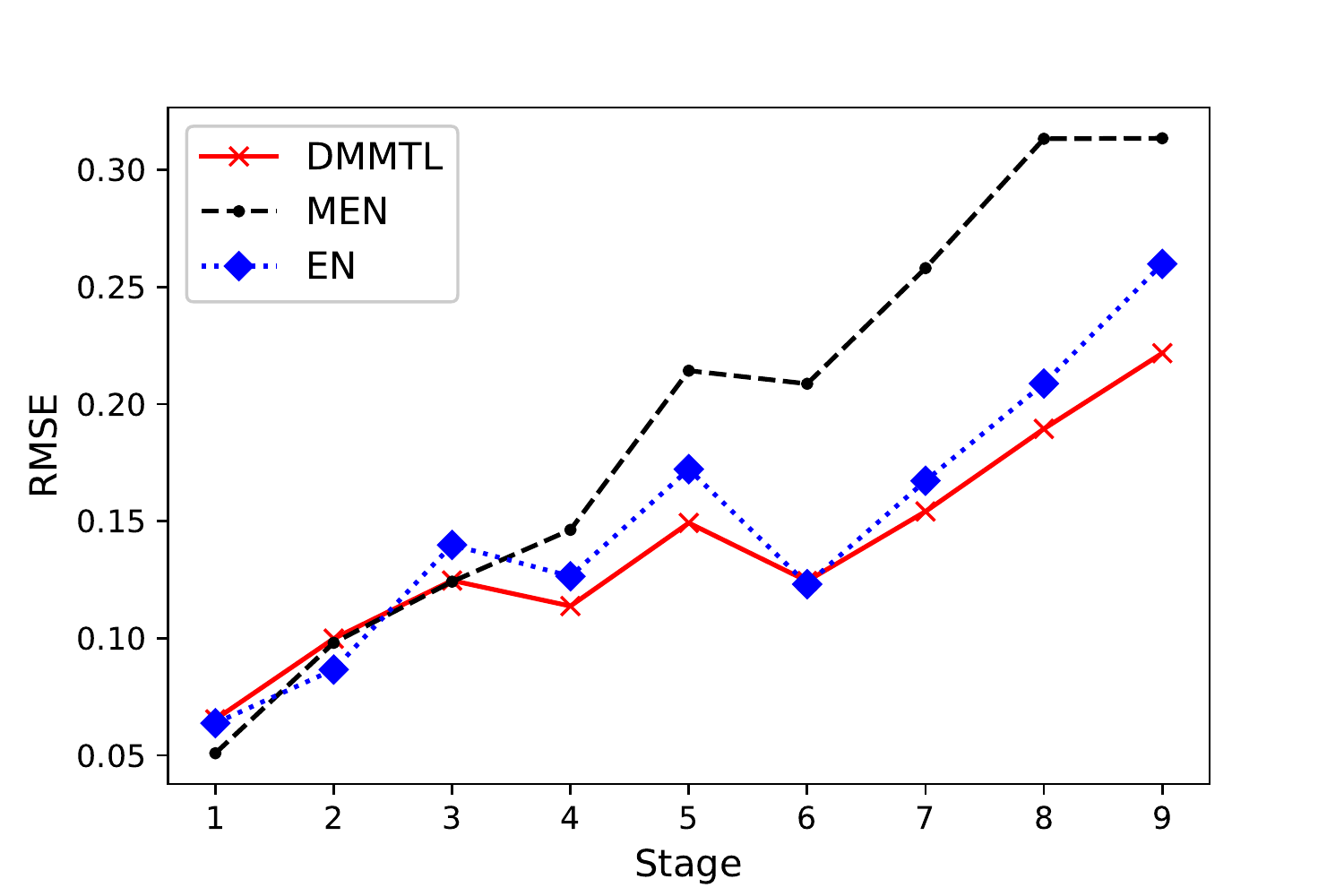}

\label{Fig: RMSEcase2}

}\subfloat[Case 3]{\includegraphics[width=0.33\linewidth]{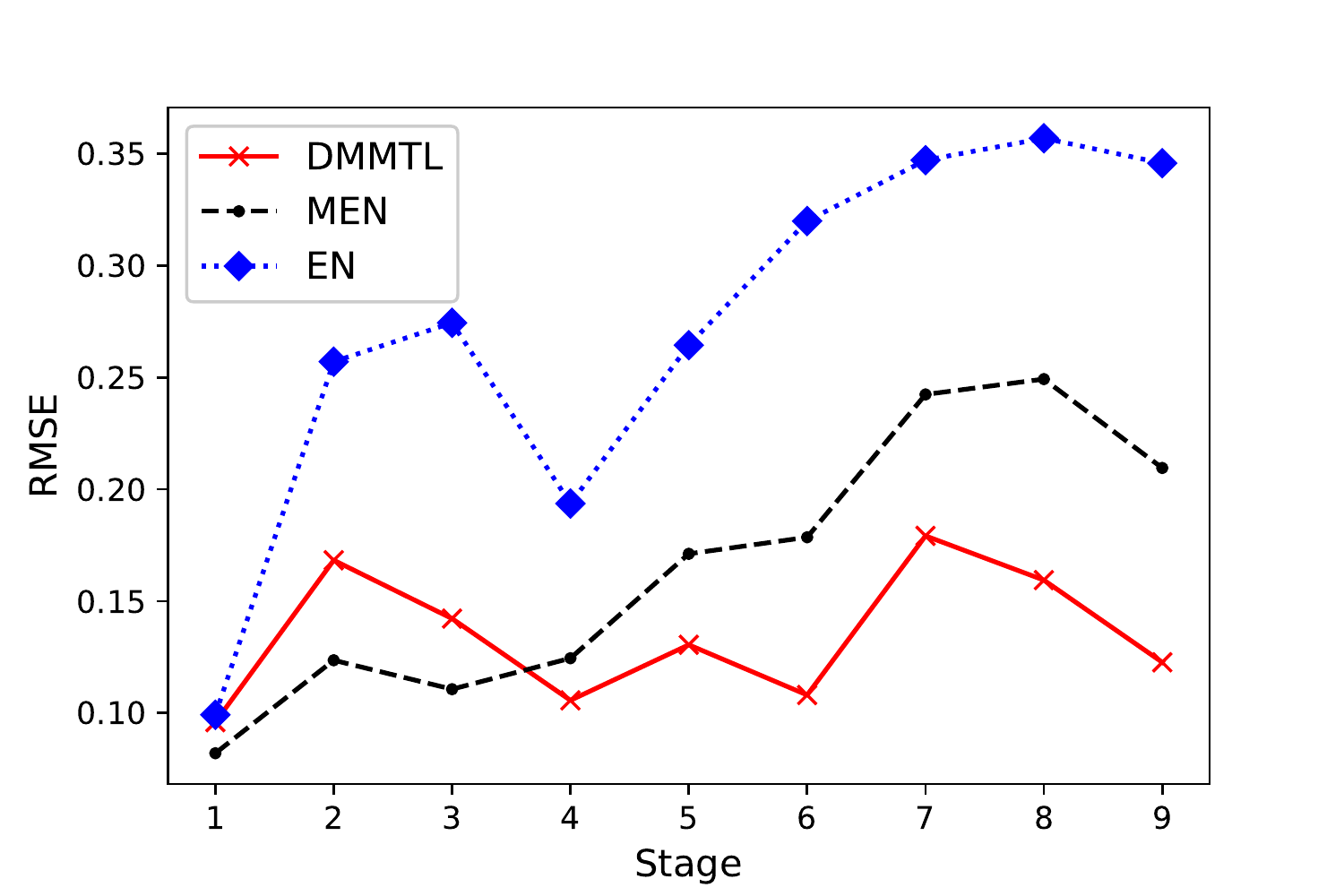}

\label{Fig: RMSEcase3}

}

\caption{RMSE according to each stage}

\label{Fig: RMSEstage}
\end{figure}

To understand how the model works, we also plot the RMSE of the top
3 methods, namely EN, MEN and DMMTL in \Cref{Fig: RMSEstage}. We
can conclude that in Case 1, DMMTL is able to keep a consistent RMSE
over all manufacturing stages due to the information provided by the
output variables in the intermediate stages. However, EN and MEN normally
demonstrate an increasing trend of the RMSE over stages. The reason
for this is that for later stages, the number of input variables increases
dramatically. Without the guidance of the output variables in the
intermediate stages, both EN and MEN cannot find the important variables
easily. In Case 2, the RMSEs of all methods increase over the stages
due to the decrease in dependency between the manufacturing stages.
However, DMMTL still outperforms others. In Case 3, DMMTL has a slightly
larger error compared to MEN in the initial stages. The reason for
this is that the first three stages in Case 3 are actually completely
independent. However, DMMTL is forced to learn a dependency between
these stages, which could lead to a worse result. When the stages
become dependent after Stage 3 (e.g., stage 4 is related to stage
1, and stage 5 is related to stage 2), DMMTL is able to quickly exploit
this dependency and outperform all other benchmark methods.

Furthermore, we would like to compare the input variable selection
accuracy for Case 2 and Case 3 for the last output variable in the
last stage (Stage 9). It is worth noting that only the proposed method,
EN and MEN, are able to perform the feature selection by selecting
the non-zero elements of the model due to the sparsity penalty. Other
benchmark models, such as LR, RF, and MLP cannot perform variable
selection naively. Recall that the data are already normalized to
mean $0$ and variance $1$. For LR, we will use the absolute value
of the model coefficient directly for the variable importance score.
For MLP, we will use only the norm of the model coefficient of the
first layer (connected to the input variables) as the input variable
importance score. For RF, we propose to use the feature importance
metric computed by the average accuracy gain of each split according
to each individual variable. The result of the percentage of the input
variable identification is shown in \Cref{Table: sensorpercentage}.

\begin{figure}
\subfloat[Case 2]{\includegraphics[width=0.5\linewidth]{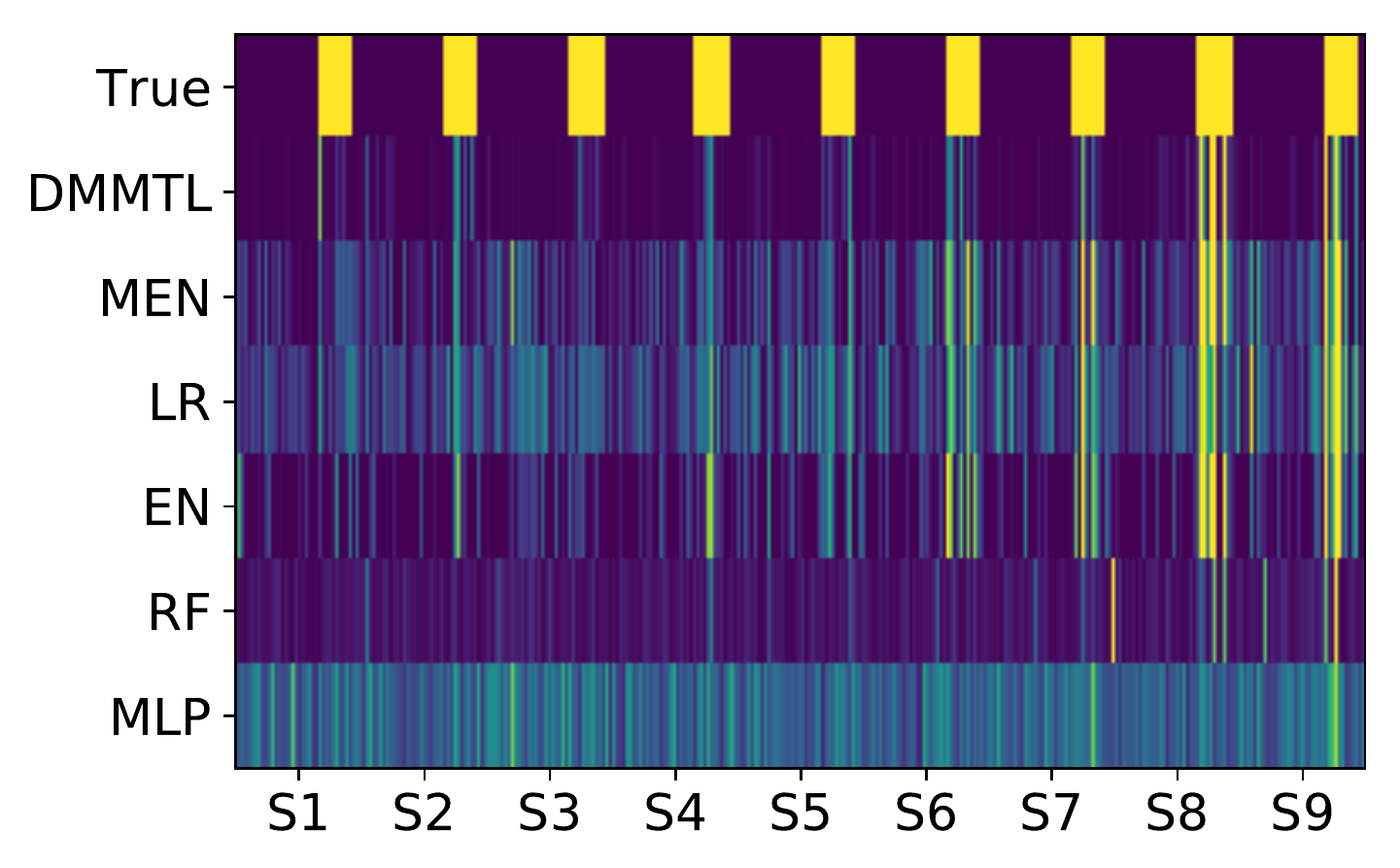}

\label{Fig: Sensorcase2}

}\subfloat[Case 3]{\includegraphics[width=0.5\linewidth]{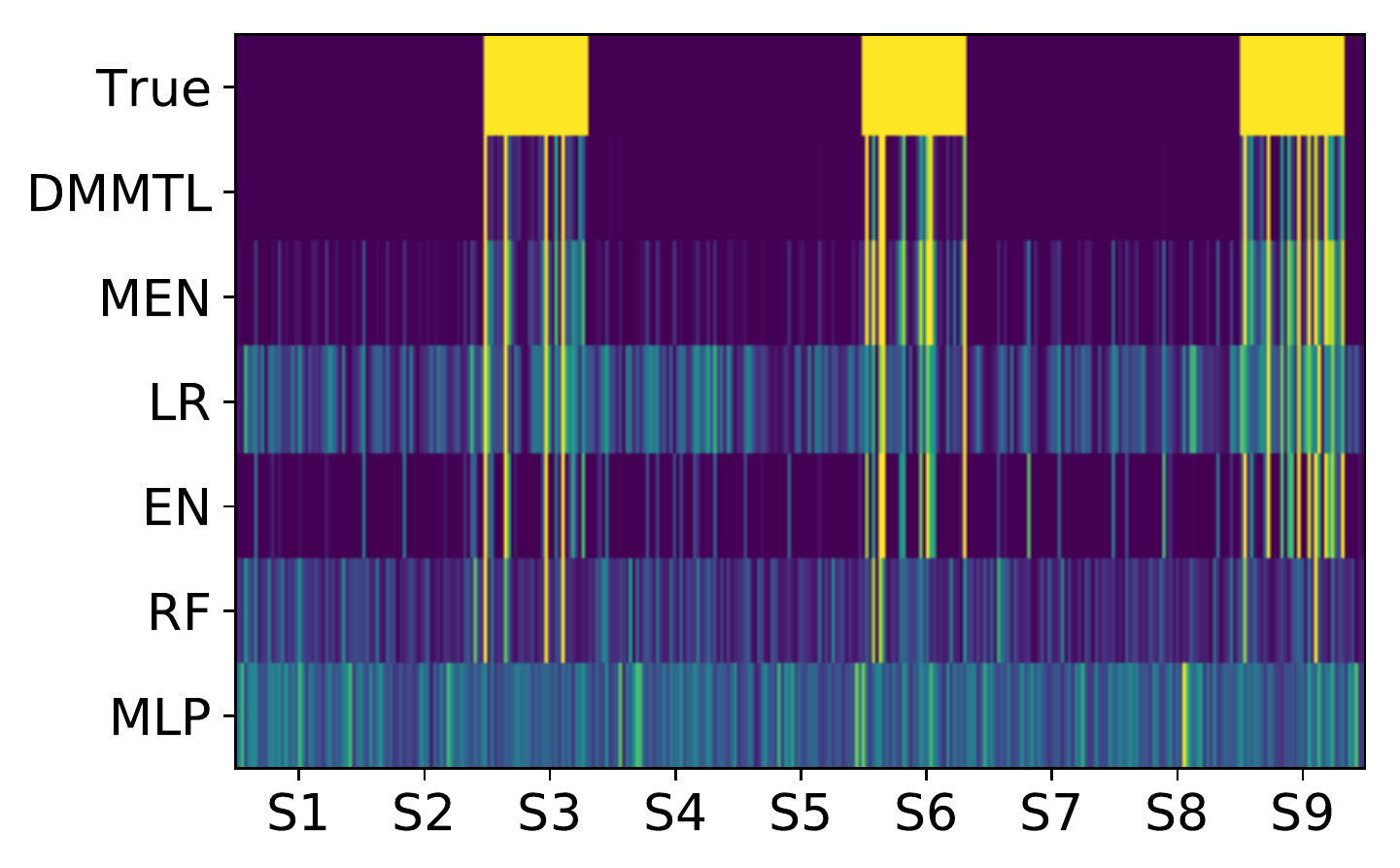}

\label{Fig:Sensorcase3}

}

\caption{Identified Important Sensor for Case 2 and Case 3}

\label{Fig: Sensor}
\end{figure}

To evaluate the variable selection accuracy, we can view the input
variable identification problem as the classification problem and
we also compute the precision, recall, and AUC score in \Cref{Table:
sensorpercentage}. Precision is defined as the percentage of identified
variables that are actually important. Recall is defined as the percentage
of important variables that are actually identified. AUC is defined
as the area under the receiver operating characteristic curve. The
threshold to determine the important input variable is set to maintain
the false positive rate as $5\%$. From \Cref{Table: sensorpercentage},
we can conclude that DMMTL is able to accurately identify the input
variables compared to other benchmark methods in both Case 2 and 3
with the highest precision, recall, and AUC score. MEN performs the
second best, due to the ability to use information from multiple output
sensors jointly within each stage. MEN is especially able to achieve
a much higher recall score than the other benchmark methods, showing
the strength of a multi-task learning framework. To better understand
how each method performs feature selection, we also plot the feature
importance score computed by each method in \Cref{Fig: Sensor}.
From \Cref{Fig: Sensor}, we can conclude that DMMTL is able to
use the least number of input variables to achieve the best prediction
power compared to all other benchmark methods due to the group lasso
penalty.

\begin{table}
\caption{Input Variable Identification Accuracy}
\centering %
\begin{tabular}{|c|cc|cc|cc|}
\hline 
 & \multicolumn{2}{c|}{Precision} & \multicolumn{2}{c|}{Recall} & \multicolumn{2}{c|}{AUC}\tabularnewline
\hline 
 & Case 2 & Case 3 & Case 2 & Case 3 & Case 2 & Case 3\tabularnewline
\hline 
DMMTL & \textbf{0.795} & \textbf{0.867} & \textbf{0.515} & \textbf{0.871} & \textbf{0.810} & \textbf{0.958}\tabularnewline
\hline 
MEN & 0.677 & 0.834 & 0.280 & 0.671 & 0.633 & 0.916\tabularnewline
\hline 
LR & 0.143 & 0.724 & 0.022 & 0.3511 & 0.486 & 0.689\tabularnewline
\hline 
EN & 0.189 & 0.752 & 0.031 & 0.404 & 0.465 & 0.706\tabularnewline
\hline 
RF & 0.167 & 0.589 & 0.027 & 0.191 & 0.463 & 0.599\tabularnewline
\hline 
MLP & 0.333 & 0.348 & 0.071 & 0.071 & 0.503 & 0.526\tabularnewline
\hline 
\end{tabular}

\label{Table: sensorpercentage}
\end{table}

\section{Case Study \label{sec:Case-Study}}

In this case study, we apply DMMTL to model the diaper assembly process
introduced in \Cref{sec: Introduction}. We divide the whole converting
process into five stages and identify $484$ process variables (e.g.
temperature, pressure, etc.) as inputs and $200$ quality measurements
(e.g. product dimensions) as output variables in the model. Due to
the complex physical process involved, it is very hard to derive the
physics relationship between the input variables and output variables.
The detailed information of the number of input and output sensors
in each stage is shown in \Cref{Table: casestudysensor}. Due to
the privacy constraint, the name of the stages and the name of the
sensors can not be given here. Furthermore, to increase the prediction
power of our model, we also use the output measurements from the previous
stage as input variables to the next stage. Because the manufacturing
data is very noisy, we use both the Huber loss function and the traditional
residual sum of squares for comparison. We also compare DMMTL with
several benchmark methods, including the multi-task elastic net (MEN),
ridge regression (RR), elastic net (EN), random forest (RF), and multi-layer
perception (MLP). We do not include linear regression (LR) in the
comparison because its parameter estimation is numerically unstable
and it can also be seen as a special case of RR without adding penalties.

\begin{table}
\caption{Number of Input and Output Variables for Each Stage}

\centering%
\begin{tabular}{|c|c|c|c|c|c|}
\hline 
\multirow{1}{*}{} & Stage 1 & Stage 2 & Stage 3 & Stage 4 & Stage 5\tabularnewline
\hline 
\hline 
Input Variables & 110 & 86 & 165 & 120 & 0\tabularnewline
\hline 
Output Variables & 20 & 64 & 10 & 90 & 16\tabularnewline
\hline 
\end{tabular}

\label{Table: casestudysensor}
\end{table}

One interesting phenomenon in the realistic case study is that not
all output variables can be predicted well based on the input variables.
Given the complexity of the manufacturing process, even with $484$
input variables some of the important characteristics of the underlying
process are still not measured by the sensors. Therefore, we try to
find a model that can achieve excellent predictive power for most
of the output variables.  Furthermore, even knowing which output variables
cannot be predicted is useful information. This could guide adding
more sensors or increasing sample frequency. Finally, for more complicated
cases, we used two-layer neural networks for both emission and transition
functions. 

\begin{figure}
\centering\includegraphics[width=0.8\linewidth]{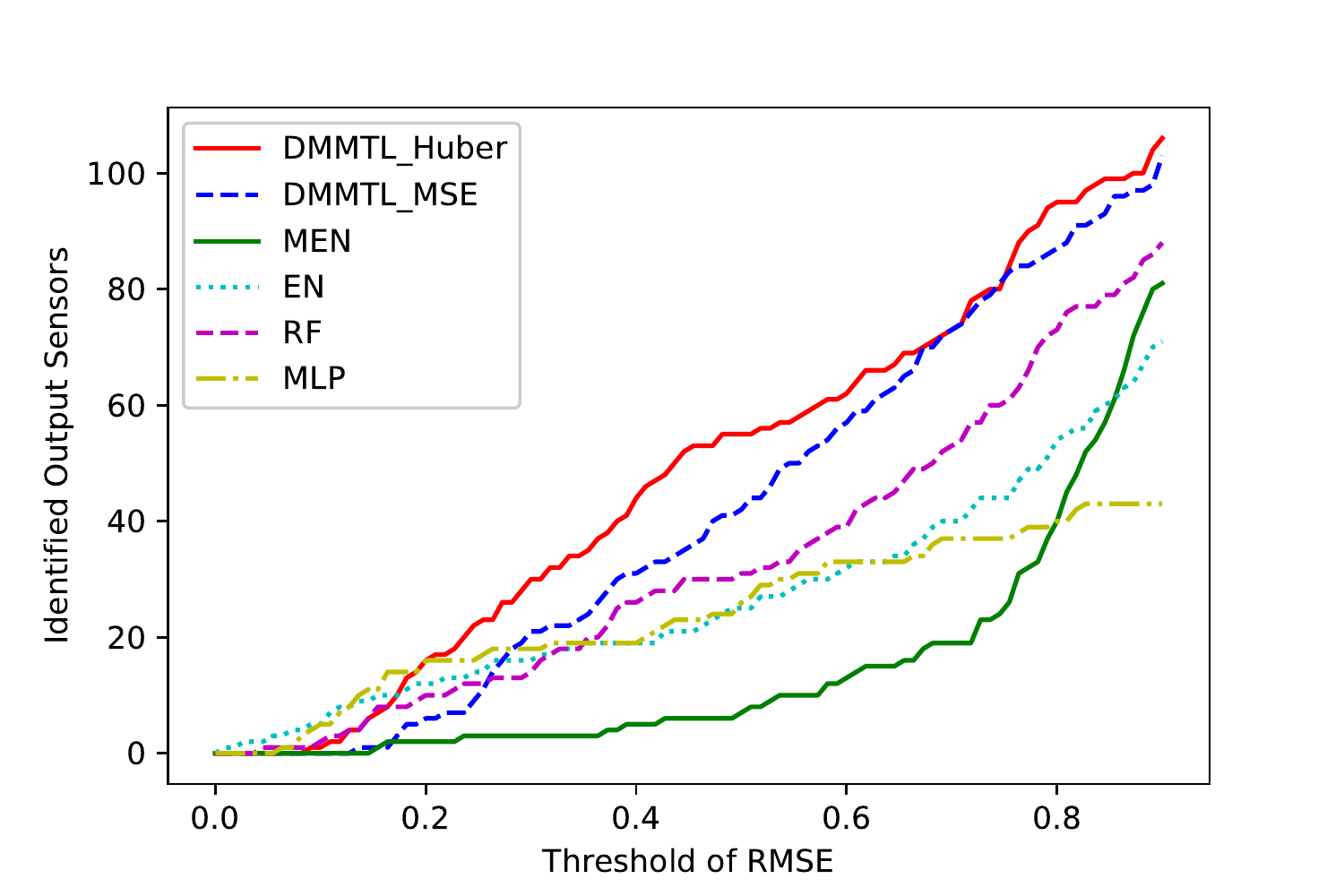}

\caption{Number of identified important output variables with different threshold
on the prediction RMSE}

\label{Fig: ImportantSensorCount}
\end{figure}

To compare how the methods are able to identify related output variables,
we first compute the RMSE of all $200$ output variables. Recall that
the RMSE is defined as $\sum_{k}\sum_{j}\|y_{k,j}^{te}-\hat{y}_{k,j}^{te}\|^{2}/\|y_{k,j}^{te}-\bar{y}_{k,j}^{tr}\|^{2}$
and if its value is smaller than $1$ then that indicates the model
can achieve a better prediction than the naive predictor based on
the mean response value. To evaluate the performance of different
methods for identifying the important variables, we first compute
the $20\%,40\%,50\%$ and $70\%$ quantiles of the $200$ RMSE scores
in \Cref{Table: casestudy} for $200$ output variables for each
method. We then choose thresholds on the RMSE ranging from $0.05$
to $0.95$ and define the number of output variables with the RMSE
score smaller than the threshold as the number of identified related
output variables. In \Cref{Fig: ImportantSensorCount}, we plot
the number of identified related output variables for different thresholds.

\Cref{Table: casestudy} shows that DMMTL can achieve a lower RMSE
over all quantiles $20\%,40\%,50\%$ and $70\%$. In particular, for
the $20\%$ quantile, DMMTL with Huber and mean square error loss
function achieve the RMSE of $0.29$ and $0.39$, which indicates
the strong prediction power (i.e., much smaller than 1). Furthermore,
when comparing the median of the RMSE, only the proposed methods are
able to achieve RMSE lower than $0.9$. This can also be seen from
\Cref{Fig: ImportantSensorCount} that DMMTL outperforms all other
benchmark methods due to the ability to combine the modeling with
multiple output variables in different stages in a unified model.
Furthermore, from \Cref{Table: casestudy}, we see that DMMTL with
the Huber loss function is able to outperform the mean square error
loss function at the $20\%$ and $40\%$ quantiles or when the threshold
is small as shown in \Cref{Fig: ImportantSensorCount}. The reason
is that the Huber loss function is more robust to outlier sensing
variables and therefore, will focus more on reducing the loss functions
of the output variables which are truly correlated to the input variables.
RF and EN's performances follow immediately after our proposed method
because they have the ability to select important variables. MLP,
in general, performs worse due to the lack of penalization and feature
selection. MEN performs the worst in this example since there are
many uncorrelated output variables even in the same stage, which violates
the assumption of MEN that the sparsity structure for all input variables
in the same stage is the same.

\begin{table}
\caption{Quantiles of Prediction RMSE}

\centering%
\begin{tabular}{|c|c|c|c|c|}
\hline 
\multirow{1}{*}{Quantile} & 20\% & 40\% & 50\% & 70\%\tabularnewline
\hline 
\hline 
DMMTL (Huber) & \textbf{0.29} & \textbf{0.73} & 0.87 & 1.01\tabularnewline
\hline 
DMMTL (MSE) & 0.39 & 0.71 & \textbf{0.81} & \textbf{0.99}\tabularnewline
\hline 
Multi-task Elastic Net & 0.79 & 0.88 & 0.92 & 0.99\tabularnewline
\hline 
Elastic Net & 0.60 & 0.91 & 0.99 & 1.00\tabularnewline
\hline 
Random Forest & 0.53 & 0.79 & 0.93 & 1.11\tabularnewline
\hline 
Multi-layer Perception & 0.76 & 1.83 & 2.59 & 5.68\tabularnewline
\hline 
\end{tabular}

\label{Table: casestudy}
\end{table}

\begin{figure}
\includegraphics[width=1\linewidth]{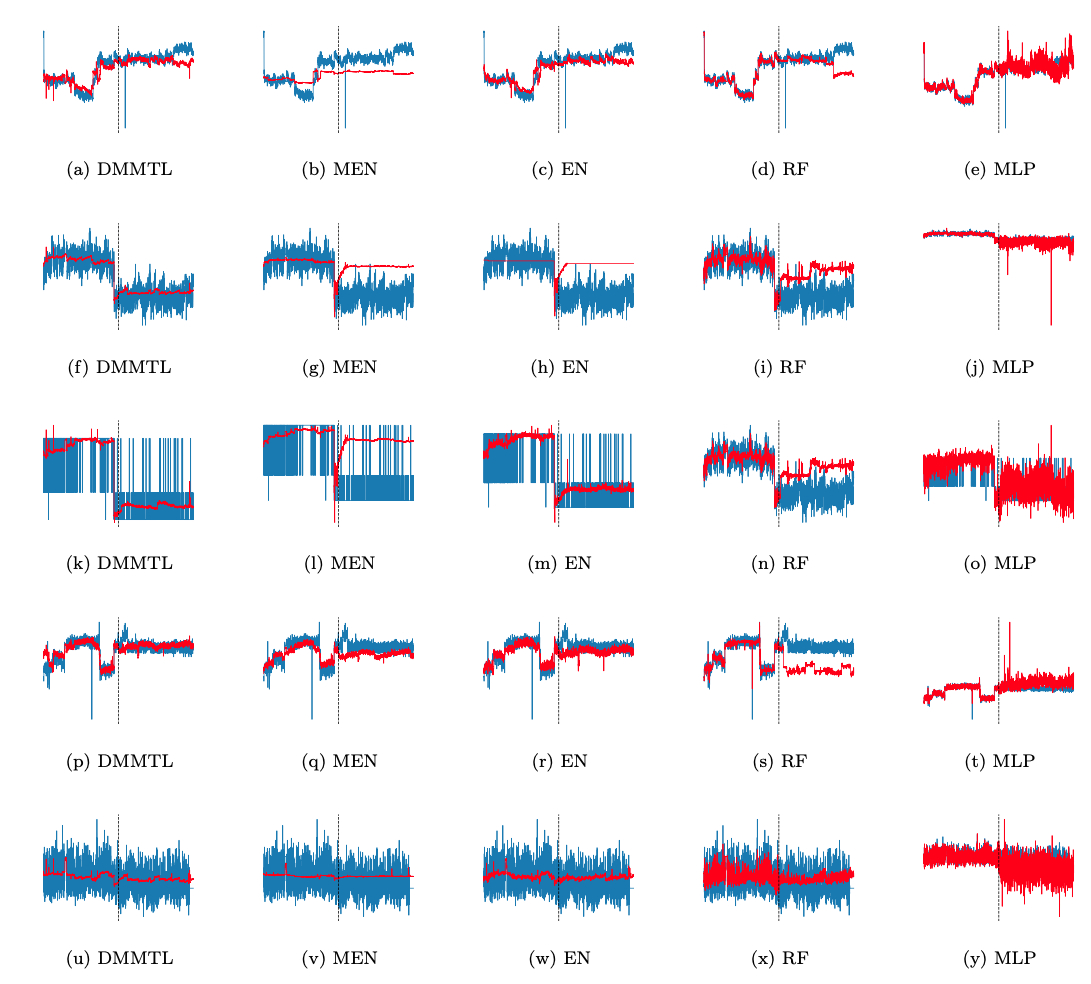}\caption{Example of Sensor Prediction (1st row: sensor 16, 2nd row: sensor
33, 3rd row: sensor 75, 4th row: sensor 132, 5th row: sensor 197)}

\label{Fig: predictsignal}
\end{figure}

To demonstrate the performance of the prediction accuracy for all
methods, in \Cref{Fig: predictsignal} we plot the predicted signals
and the true signal for the output variables for both training and
testing data. A dashed black line is added in each plot to separate
the training and testing data. We select output sensors 16, 33, 75,
132, 197 for demonstration in stage 1, 2, 3, 4, 5, respectively. The
RMSE according to these output sensors and the selected number of
input variables of each method are shown in \Cref{Table: IdentifiedNumber}
and \Cref{Table: RMSE}.

From \Cref{Fig: predictsignal} and \Cref{Table: IdentifiedNumber},
we first conclude that these output variables share similar patterns.
For example, Output 16 has a meanshift during time $1961$ and Output
33, 75, and 197 has a meanshift at time 2813. Output 132 have meanshifts
at both time points. DMMTL is able to achieve the least RMSE among
all methods due to its ability to combine all output sensors in a
unified model, therefore leading to a better model for all output
sensors. EN also achieves good performance for Output 16 and 75. For
Output 33 and 132, only DMMTL is able to accurately predict the trend.
RF sometimes does not capture the trend correctly. MLP typically overfits
the data, and therefore it normally produces much larger noises in
the testing data. MEN typically underfits the data due to its strong
assumption that the models for output sensors in the same stage must
share the same sparsity patterns. In terms of the number of input
sensors identified, typically EN is able to identify the least number
of input sensors, followed by DMMTL and MEN. MLP is not able to perform
feature selection, which leads to severe over-fitting. In conclusion,
we see that DMMTL is able to achieve the least RMSE with a relatively
small number of selected input sensors.

\begin{table}
\caption{Number of Input Sensors Identified}

\centering%
\begin{tabular}{|c|ccccc|}
\hline 
Sensor \# & 16 & 33 & 75 & 132 & 197\tabularnewline
\hline 
DMMTL & 16 & 20 & 21 & 21 & 21\tabularnewline
\hline 
MEN & 35 & 35 & 35 & 35 & 36\tabularnewline
\hline 
EN & 18 & 18 & 18 & 16 & 16\tabularnewline
\hline 
RF & 21 & 116 & 151 & 30 & 342\tabularnewline
\hline 
MLP & 119 & 225 & 225 & 593 & 682\tabularnewline
\hline 
\end{tabular}

\label{Table: IdentifiedNumber}
\end{table}

\begin{table}
\caption{RMSE}
\centering %
\begin{tabular}{|c|ccccc|}
\hline 
Sensor \# & 16 & 33 & 75 & 132 & 197\tabularnewline
\hline 
DMMTL & \textbf{0.12} & \textbf{0.07} & \textbf{0.05} & \textbf{0.37} & \textbf{0.73}\tabularnewline
\hline 
MEN & 0.59 & 0.76 & 0.83 & 1.21 & 0.95\tabularnewline
\hline 
EN & 1.00 & 0.89 & 0.09 & 1.00 & 0.98\tabularnewline
\hline 
RF & 0.39 & 0.65 & 0.41 & 7.73 & 0.83\tabularnewline
\hline 
MLP & 0.16 & 3.94 & 0.32 & 2.84 & 7.33\tabularnewline
\hline 
\end{tabular}\label{Table: RMSE}
\end{table}

\begin{figure}
\includegraphics[width=1\linewidth]{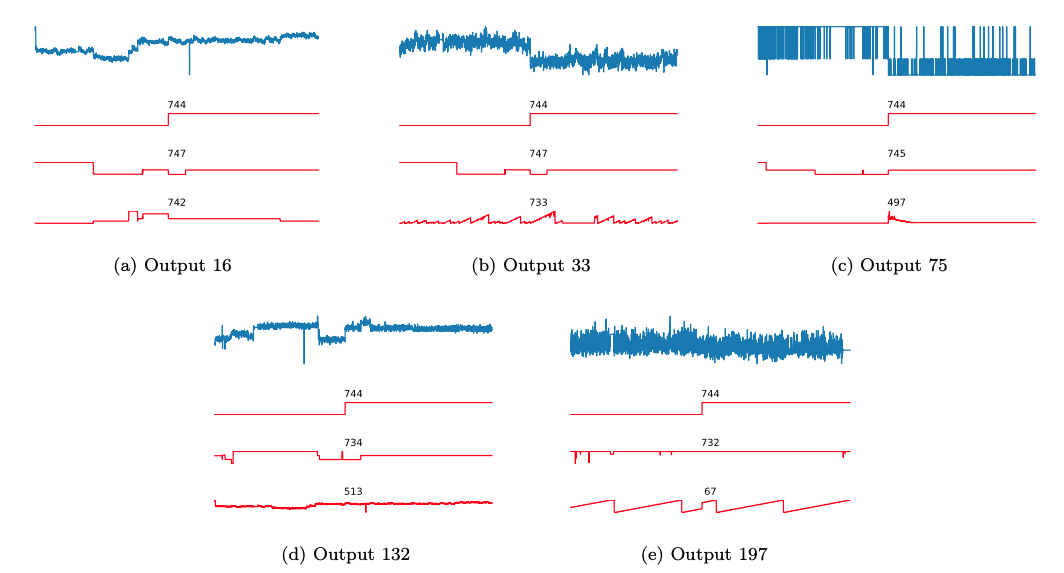}

\caption{Top three important input variables identified by DMMTL for Output
16, 33, 75, 132 and 197}

\label{Fig: Identified}
\end{figure}

Finally, in \Cref{Fig: Identified}, we also plot the top three
important input variables identified for all these five output variables.
Among them, Input 744 can explain the meanshift at time $2813$ for
all output variables and Input 747 can explain some of the other small
meanshifts for Output 16 and 33. Finally, Input 733 and Input 67 are
able to explain the small increasing trend in Output 33 and 197, respectively.
We have validated that the selected input variables indeed can explain
the variation for the selected output variable from the domain knowledge.

\section{Conclusion \label{sec:Conclusion}}

Modeling complex multistage manufacturing systems is an important
research topic for accurate process prediction, monitoring, diagnosis
and control. This paper proposes a deep transition model with multi-task
learning to jointly model all output sensing variables with the input
sensing variables according to the sequential production line structure.
Furthermore, since the dimensionality of the input sensing variables
and output sensing variables can be very high, we suggest reducing
the dimensionality by utilizing the sparse regularization and robust
Huber loss function to select the important sensing variables. DMMTL
has been tested through several simulated studies and a realistic
case study of a real diaper manufacturing system. The results demonstrate
that it achieves a better prediction accuracy as well as a better
local and global interpretability by identifying important relationship
between input and output sensing variables.

There are several future research directions that we would like to
investigate. One is to extend this method to heterogeneous measurements
with more stage dependencies (e.g., tree structures) other than the
production line set up in series. Another extension is to study the
proposed algorithm under the stochastic transition of the hidden variables
similar to the stream of variation models.

\appendix

\section{Relationship to Stream of Variation}

\label{sec: sov} The foundation of SoV methodology is a mathematical
model that links the output variables (e.g., key quality characteristics
of the product ) with key input variables (e.g., key process sensing
variables) through the state space representation. 
\begin{equation}
\mathbf{h}_{k}=\mathbf{W}_{xk}\mathbf{x}_{k}+\mathbf{U}_{hk}\mathbf{h}_{k-1}+\mathbf{e}_{hk},\mathbf{y}_{k}=\mathbf{V}_{yk}\mathbf{h}_{k}+\mathbf{e}_{yk}\label{eq: Transition}
\end{equation}
The variable $\mathbf{h}_{k}$ is the state vector representing the
output variables at stage $k$. $\mathbf{e}_{hk}$ and $\mathbf{e}_{yk}$
are the modeling error and measurement error, respectively. The coefficient
matrices $\mathbf{W}_{xk}$, $\mathbf{U}_{hk}$, and $\mathbf{V}_{yk}$
are determined by product and process design information at stage
$k$. $\mathbf{W}_{xk}$ represents the impact of the new stage process
to the product. $\mathbf{U}_{hk}$ represents the transition from
stage $k-1$ to stage $k$. $\mathbf{V}_{yk}$ is the measurement
matrix, which links the hidden state $\mathbf{h}_{k}$ and the output
$\mathbf{y}_{k}$.

These mathematical models have achieved great success in MMS modeling
by integrating the product and process design information and modeling
the variation propagation in the MMS. However, the SoV methodology
assumes that the key output variables $\mathbf{y}_{k}$ and key input
variables $\mathbf{x}_{k}$ have been correctly identified. Furthermore,
it requires the matrices $\mathbf{W}_{xk}$, $\mathbf{U}_{hk}$, and
$\mathbf{V}_{yk}$ are known in each stage $k$, which is not possible
if the system is too complex. Finally, it assumes the linear transition
matrix between states which could be an over-simplification in many
real cases. However, SoV assumes that the transition between the state
variables are known as the linear stochastic function. The proposed
DMMTL assumes that the transition is unknown nonlinear functions.
Extending the framework to stochastic functions would be one of our
future work.

\section{Relationship to Recurrent Neural Network \label{subsec: RNN}}

The formulation of RNN methodology is a neural network that links
the output variables $\mathbf{y}_{k}$, input variables $\mathbf{x}_{k}$
in a neural network. More specifically, at each time $k$, the output
$\mathbf{y}_{k}$ and input $\mathbf{x}_{k}$ are linked together
via the hidden state $\mathbf{h}_{k}$ in (\Cref{eq: RNN}). 
\begin{equation}
\mathbf{h}_{k}=\sigma(\mathbf{W}_{x}\mathbf{x}_{k}+\mathbf{U}_{h}\mathbf{h}_{k-1}+\mathbf{b}_{h}),\mathbf{y}_{k}=\mathbf{V}_{y}\mathbf{h}_{k}+\mathbf{b}_{y}\label{eq: RNN}
\end{equation}
In RNN, it is typically assumed that the system is time-invariant,
which means the model parameters $\mathbf{W}_{x},\mathbf{U}_{x},\mathbf{V}_{y}$
are independent of $k$. RNN has achieved great success in machine
translation problems \citep{Graves}, handwriting recognition \citep{graves2009offline}
and speech recognition \citep{graves2013speech}. Furthermore, model
parameters $\mathbf{W}_{x},\mathbf{U}_{x},\mathbf{V}_{y}$ can be
learned in an end-to-end fashion via the combination of back-propagation
\citep{lecun1990handwritten} and stochastic gradient descent \citep{bottou2010large}.

However, the major limitation of using RNN in MMS is that different
manufacturing stages are inherently different. The underlying physics
is entirely different for each stage which not only results in the
different transition matrix $\mathbf{W}_{x}$, $\mathbf{U}_{h}$,
and $\mathbf{V}_{y}$. RNN also assumes the same set of variables
are predicted in each time. However, in MMS, different quality inspection
sensors are set up in each manufacturing stage. Finally, RNN is a
complicated model and can not achieve input and output variable selection
as the proposed approach. 

\section{Proof of Proposition 2: \label{sec:proximalGrad}}
\begin{proof}
Considering the loss function 
\begin{align}
\min_{a_{kj}}\sum_{k,j}\|y_{kj}-g_{kj}(\mathbf{h}_{k};\boldsymbol{\Theta})-a_{kj}\|^{2}+\lambda_{x}\sum_{i=1}^{n_{x,k}}\|\mathbf{w}_{i,xk}\|_{2}+\frac{\lambda}{2}\|\Theta\|^{2}+\gamma\sum_{k,j}\|a_{kj}\|_{1}\label{eq: lrobustloss}
\end{align}
Here, the loss function in (\ref{eq: lrobustloss}) can be decoupled
into each pair of $(k,j)$ individually. Therefore, each $a_{kj}$
can be solved individually by optimizing 
\[
\|y_{kj}-g_{kj}(\mathbf{h}_{k};\boldsymbol{\Theta})-a_{kj}\|^{2}+\gamma\|a_{kj}\|_{1}
\]
and can be solved by 
\[
a_{kj}=S_{\gamma/2}(y_{kj}-g_{kj}(\mathbf{h}_{k};\boldsymbol{\Theta})),
\]
To optimize the $\mathbf{w}_{i,xk}$, we will follow the derivation
of the proximal gradient algorithm the Taylor expansion of $\mathcal{L}(\Theta,\mathcal{A})$
as
\begin{equation}
\mathcal{L}(\Theta,\mathcal{A})\leq\mathcal{L}(\Theta^{(t-1)},\mathcal{A})+\sum_{i}\frac{\partial\mathcal{L}(\Theta^{(t-1)},\mathcal{A})}{\partial\mathbf{w}_{i,xk}}\left(\mathbf{w}_{i,xk}-\mathbf{w}_{i,xk}^{(t-1)}\right)+\frac{L}{2}\|\mathbf{w}_{i,xk}-\mathbf{w}_{i,xk}^{(t-1)}\|^{2}\label{eq: RHS}
\end{equation}
$L$ is the Lipschitz constance of $\mathcal{L}(\Theta,\mathcal{A})$.
Therefore, we aim to minimize the upper bound of $\mathcal{L}(\Theta,\mathcal{A})+\mathcal{R}(\Theta,\mathcal{A})$
as follows: 
\begin{align*}
\min_{\{\mathbf{w}_{i,xk}\}}\mathcal{L}(\Theta^{(t-1)},\mathcal{A})+\sum_{i,k}\frac{\partial\mathcal{L}(\Theta^{(t-1)},\mathcal{A})}{\partial\mathbf{w}_{i,xk}}\left(\mathbf{w}_{i,xk}-\mathbf{w}_{i,xk}^{(t-1)}\right)+\frac{L}{2}\sum_{i,k}\|\mathbf{w}_{i,xk}-\mathbf{w}_{i,xk}^{(t-1)}\|^{2}+\\
\lambda_{x}\sum_{i,k}\|\mathbf{w}_{i,xk}\|_{2}+\frac{\lambda}{2}\sum_{i,k}\|\mathbf{w}_{i,xk}\|^{2}
\end{align*}
Here, we find that the optimization can be decoupled to each individual
$(i,k)$ as 
\[
\min_{\mathbf{w}_{i,xk}}\|\mathbf{w}_{i,xk}-\frac{L}{L+\lambda}(\mathbf{w}_{i,xk}^{(t-1)}-\frac{1}{L}\frac{\partial\mathcal{L}(\Theta^{(t-1)},\mathcal{A})}{\partial\mathbf{w}_{i,xk}})\|^{2}+\frac{2\lambda_{x}}{L+\lambda}\sum_{i=1}^{n_{x,k}}\|\mathbf{w}_{i,xk}\|_{2},
\]
and can be solved in closed form as
\[
\mathbf{w}_{i,xk}^{(t)}=S_{\frac{\lambda_{x}}{L+\lambda}}(\frac{L}{L+\lambda}(\mathbf{w}_{i,xk}^{(t-1)}-\frac{1}{L}\frac{\partial\mathcal{L}(\Theta^{(t-1)},\mathcal{A})}{\partial\mathbf{w}_{i,xk}})).
\]
\end{proof}

\section{Back-propagation along the Sequential Stages Over Parameters $\boldsymbol{\theta}_{k}$
\label{sec:Gradient}}

We will discuss how to efficiently optimize the model parameters $\Theta$
via stochastic gradient descent. Suppose we denote $\mathcal{X}{}^{n},\mathcal{Y}^{n}$
as the $n^{th}$ sample of the entire dataset with $n=1,\cdots,N_{n}$,
then the loss function can be decomposed as 
\begin{equation}
\mathcal{L}(\mathbf{\Theta};\{\mathcal{X}\},\{\mathcal{Y}\})=\frac{1}{N_{n}}\sum_{n=1}^{N_{n}}\sum_{k=1}^{K}\sum_{j=1}^{N_{y,k}}\log P(y_{kj}^{n}|\mathbf{h}_{k};\Theta).
\end{equation}
If the number of samples $N_{n}$ is large, averaging the gradient
over the entire dataset is normally slow. To address this, we propose
to apply the mini-batch stochastic gradient algorithm, which is widely
used to optimize large-scale machine learning problems. In each iteration
$t$, we can choose a subset of samples $\mathcal{N}_{t}\in\{1,\cdots,N_{n}\}$,
where the gradient is only evaluated as the average of the subset
of the entire samples in (\ref{eq: sgd}): 
\begin{equation}
\boldsymbol{\mathbf{\Theta}}^{(t+1)}=\boldsymbol{\mathbf{\Theta}}^{(t)}-\frac{1}{|\mathcal{N}_{t}|}\sum_{n\in\mathcal{N}_{t}}\frac{\partial\tilde{L}(\mathcal{X}{}^{n},\mathcal{Y}^{n};\mathbf{\mathbf{\Theta}}^{(t)})}{\partial\mathbf{\Theta}}.\label{eq: sgd}
\end{equation}

Now we will discuss how to compute the gradient according to the model
coefficients $\mathbf{\Theta}$. $\mathbf{\Theta}=\{\boldsymbol{\theta}_{1},\cdots\boldsymbol{\theta}_{K}\}$.
First, the likelihood can be decomposed into different samples, stages,
and output variables due to the conditional dependency of the hidden
variables $\mathbf{h}_{k'}$ in (\ref{eq: decomploss}). 
\begin{equation}
\frac{\partial}{\partial\boldsymbol{\theta}_{k}}\mathcal{L}(\mathbf{x},\mathbf{y};\Theta)=\sum_{n=1}^{N_{n}}\sum_{k'=1}^{K}\sum_{j=1}^{n_{y,k'}}\frac{\partial}{\partial\boldsymbol{\theta}_{k}}\log P(y_{k'j}^{n}|\mathbf{h}_{k'};\boldsymbol{\theta}_{k}^{g})\label{eq: decomploss}
\end{equation}
Therefore, we need to compute $\frac{\partial}{\partial\boldsymbol{\theta}_{k}}\log P(y_{k'j}^{n}|\mathbf{h}_{k'};\boldsymbol{\theta}_{k}^{g})$.
For $k'<k$, it is obvious that \\
 $\frac{\partial}{\partial\boldsymbol{\theta}_{k}}\log P(y_{k'j}^{n}|\mathbf{h}_{k'};\boldsymbol{\theta}_{k}^{g})=0$.
However, if $k'\geq k$, we can compute the gradient $\frac{\partial\log P(y_{k'j}^{n}|\mathbf{h}_{k'};\boldsymbol{\theta}_{k}^{g})}{\partial\boldsymbol{\theta}_{k}}$
recursively as $\frac{\partial\log P(y_{k'j}^{n}|\mathbf{h}_{k'};\boldsymbol{\theta}_{k}^{g})}{\partial\boldsymbol{\theta}_{k}}=\frac{\partial\log P(y_{k'j}^{n}|\mathbf{h}_{k'};\boldsymbol{\theta}_{k}^{g})}{\partial\mathbf{h}_{k'}}\frac{\partial\mathbf{h}_{k'}}{\partial\mathbf{h}_{k'-1}}\cdots\frac{\partial\mathbf{h}_{k+1}}{\partial\mathbf{h}_{k}}\frac{\partial\mathbf{h}_{k}}{\partial\boldsymbol{\mathbf{\theta}}_{k}}$.
By plugging in this into (\ref{eq: decomploss}), we can derive the
gradient according to the state transition parameters (\ref{eq: thetahgrad}).
\begin{equation}
\frac{\partial}{\partial\boldsymbol{\theta}_{k}^{h}}\mathcal{L}(\mathbf{x},\mathbf{y};\Theta)=\sum_{k'=k}^{K}\sum_{j=1}^{n_{y,k'}}\frac{\partial\log P(y_{k'j}^{n}|\mathbf{h}_{k'};\boldsymbol{\theta}_{k}^{g})}{\partial\mathbf{h}_{k'}}\frac{\partial\mathbf{h}_{k'}}{\partial\mathbf{h}_{k'-1}}\cdots\frac{\partial\mathbf{h}_{k+1}}{\partial\mathbf{h}_{k}}\frac{\partial\mathbf{h}_{k}}{\partial\boldsymbol{\mathbf{\theta}}_{k}^{h}}.\label{eq: thetahgrad}
\end{equation}

Finally, to derive the gradient according to output coefficient $\boldsymbol{\theta}_{k}^{g}$,
we can derive 
\begin{equation}
\frac{\partial}{\partial\boldsymbol{\theta}_{k}^{g}}\mathcal{L}(\mathbf{x},\mathbf{y};\Theta)=\sum_{j=1}^{n_{y,k}}\frac{\partial\log P(y_{kj}^{n}|\mathbf{h}_{k};\boldsymbol{\theta}_{k}^{g})}{\partial\boldsymbol{\theta}_{k}^{g}}.\label{eq: thetaggrad}
\end{equation}

We would then discuss how to compute the gradient based for $\frac{\partial\log P(y_{kj}|\mathbf{h}_{k};\boldsymbol{\theta}_{k}^{g})}{\partial\mathbf{h}_{k}}$,
$\frac{\partial\log P(y_{kj}|\mathbf{h}_{k};\boldsymbol{\theta}_{k}^{g})}{\partial\boldsymbol{\theta}_{k}^{g}}$,
$\frac{\partial\mathbf{h}_{k}}{\partial\boldsymbol{\theta}_{k}^{h}}$,
$\frac{\partial\mathbf{h}_{k}}{\partial\mathbf{h}_{k-1}}$. The computation
is shown as follows:
\begin{itemize}
\item Emission layer
\[
\frac{\partial\log P(y_{kj}|\mathbf{h}_{k};\boldsymbol{\theta}_{k}^{g})}{\partial\mathbf{h}_{k}}=2(g_{kj}(\mathbf{h}_{k};\boldsymbol{\theta}_{k}^{g})-y_{k})\frac{\partial g_{kj}(\mathbf{h}_{k};\boldsymbol{\theta}_{k}^{g})}{\partial\mathbf{h}_{k}}
\]
\[
\frac{\partial\log P(y_{kj}|\mathbf{h}_{k};\boldsymbol{\theta}_{k}^{g})}{\partial\boldsymbol{\theta}_{k}^{g}}=2(g_{kj}(\mathbf{h}_{k};\boldsymbol{\theta}_{k}^{g})-y_{k})\frac{\partial g_{kj}(\mathbf{h}_{k};\boldsymbol{\theta}_{k}^{g})}{\partial\boldsymbol{\theta}_{k}^{g}}
\]
\item Transition layer 
\[
\frac{\partial\mathbf{h}_{k}}{\partial\mathbf{h}_{k-1}}=\frac{\partial f_{k}(\mathbf{h}_{k-1},x_{k};\boldsymbol{\theta}_{k}^{h})}{\partial\mathbf{h}_{k-1}},\frac{\partial\mathbf{h}_{k}}{\partial\boldsymbol{\theta}_{k}^{h}}=\frac{\partial f_{k}(\mathbf{h}_{k-1},x_{k};\boldsymbol{\theta}_{k}^{h})}{\partial\boldsymbol{\theta}_{k}^{h}},\frac{\partial\mathbf{h}_{k}}{\partial\mathbf{x}_{k}}=\frac{\partial f_{k}(\mathbf{h}_{k-1},\mathbf{x}_{k};\boldsymbol{\theta}_{k}^{h})}{\partial\mathbf{x}_{k}}
\]
\end{itemize}
Depending on the different function forms of $f_{k}(\cdot)$ and $g_{k}(\cdot)$,
the gradient can be computed.

\section{Multi-task Elastic Net \label{sec:Gradient-1}}

In this section, we will briefly review the multi-task elastic net
methods in the literature. Here, $Y\in\mathbb{R}^{n\times k}$ is
the multivariate output, where $n$ is the number of the samples and
$k$ is the number of tasks\@. $X\in\mathbb{R}^{n\times p}$ is the
input variables and $p$ is the number of feature variables. $W\in\mathbb{R}^{p\times k}$
is the regression coefficient matrix. 

Finally, to impose the similarity between multiple tasks, the following
regularization is used as follows. 
\[
\frac{1}{2n}\|Y-XW\|^{2}+\alpha\beta\|W\|_{2,1}+\frac{1}{2}\alpha(1-\beta)\|W\|^{2}.
\]
Here, $\|W\|_{2,1}=\sum_{i}\sqrt{\sum_{j}W_{ij}^{2}}$ is the sum
of norm of each row of $W$, where $\|W\|_{2,1}$ encourages the same
sparsity patterns among all tasks. For example, it is easy to show
that when the number of tasks equals to one, the problem MEN reduces
to the $L_{1}$-norm regularized optimization problem (or Lasso).
When there are multiple tasks, the weights corresponding to the $i$-th
feature are grouped together via the $L_{2}$-norm of $W$. Furthermore,
$\|W\|^{2}$ is used to further regularize the model to avoid the
multicollinearity of the task input matrix $X$. $\alpha$ and $\beta$
are two tuning parameters control the level of regularization. 

\bibliographystyle{apalike}
\bibliography{References}

\end{document}